\documentclass[journal]{IEEEtran}

\usepackage{graphicx}
\usepackage{epstopdf}
\usepackage{subfigure}
\usepackage[numbers,sort&compress]{natbib}

\usepackage{caption2}
\usepackage{amsmath}
\usepackage{array,diagbox,siunitx}

\usepackage{cases}
\usepackage{float}
\usepackage{amssymb}
\usepackage{multirow}
\usepackage{mathrsfs}
\usepackage{algorithmic}
\usepackage[ruled]{algorithm2e}
\usepackage{rotating}
\usepackage{balance}
\usepackage{threeparttable}
\usepackage{color}
\usepackage{colortbl}
\definecolor{mygray}{gray}{.9}
\usepackage{makecell}
\usepackage{pdflscape}
\usepackage{siunitx}
\usepackage{stfloats}
\usepackage{verbatim}
\usepackage{setspace}
\usepackage{threeparttable}
\usepackage{supertabular,booktabs}
\usepackage{rotating}
\usepackage{supertabular}
\usepackage{pifont}
\usepackage{xcolor,pict2e}

\usepackage{tikz}
\usepackage{booktabs}
\usepackage{footnote}
\usepackage[pagewise]{lineno}
\usepackage{balance}
\newcommand{\etal}{\textit{et al.}}
\newcommand{\ie}{\textit{i.e.}}
\newcommand{\eg}{\textit{e.g.}}

\begin{document}
\title{An Empirical Study of Super-resolution on Low-resolution Micro-expression Recognition}
\author{Ling Zhou, 
        Mingpei Wang,
        Xiaohua Huang, ~\IEEEmembership{Senior Member,~IEEE},
        Wenming Zheng, ~\IEEEmembership{Senior Member,~IEEE},
        Qirong Mao,
        Guoying Zhao,~\IEEEmembership{Fellow,~IEEE}
\thanks{L. Zhou and M. Wang are with  the School of Computer Science and Engineering, Macau University of Science and Technology, Macau, China. (e-mail:lzhou@must.edu.mo, 2220019007@student.must.edu.mo)}

\thanks{X. Huang is with the Key Laboratory of Child Development and Learning Science (Southeast University), Ministry of Education, Southeast University, Nanjing 210096, China and also with School of Computer Engineering, Nanjing Institute of Technology, Jiangsu, China.
(email: xiaohua.huang@njit.edu.cn)}

\thanks{W. Zheng is with the Key Laboratory of Child Development and Learning Science (Southeast University), Ministry of Education, Southeast University, Nanjing 210096, China and also with the School of Biological Science and Medical Engineering, Southeast University, Nanjing 210096, Jiangsu, China.
(E-mail: wenming\_zheng@seu.edu.cn)}

\thanks{Q. Mao is with the School of Computer Science and Communication Engineering, Jiangsu University, Zhenjiang, Jiangsu, China.
(e-mail: mao\_qr@ujs.edu.cn)}

\thanks{G. Zhao is with the Center for Machine Vision and Signal Analysis, University of Oulu, Finland.
(e-mail: guoying.zhao@oulu.fi)}
}

\markboth{}%
{Shell \MakeLowercase{\textit{et al.}}: Bare Demo of IEEEtran.cls for IEEE Journals}

\maketitle
\begin{abstract}


Micro-expression recognition (MER) in low-resolution (LR) scenarios presents an important and complex challenge, particularly for practical applications such as group MER in crowded environments. Despite considerable advancements in super-resolution techniques for enhancing the quality of LR images and videos, few study has focused on investigate super-resolution for improving LR MER. The scarcity of investigation can be attributed to the inherent difficulty in capturing the subtle motions of micro-expressions, even in original-resolution MER samples, which becomes even more challenging in LR samples due to the loss of distinctive features. Furthermore, a lack of systematic benchmarking and thorough analysis of super-resolution-assisted MER methods has been noted. This paper tackles these issues by conducting a series of benchmark experiments that integrate both super-resolution (SR) and MER methods, guided by an in-depth literature survey. Specifically, we employ seven cutting-edge state-of-the-art (SOTA) MER techniques and evaluate their performance on samples generated from 13 SOTA SR techniques, thereby addressing the problem of super-resolution in MER. Through our empirical study, we uncover the primary challenges associated with SR-assisted MER and identify avenues to tackle these challenges by leveraging recent advancements in both SR and MER methodologies. Our analysis provides insights for progressing toward more efficient SR-assisted MER.

\end{abstract}

\IEEEpeerreviewmaketitle
\section{Introduction} \label{sec:Intro}
\begin{figure}
    \centering
    \includegraphics[width=\linewidth]{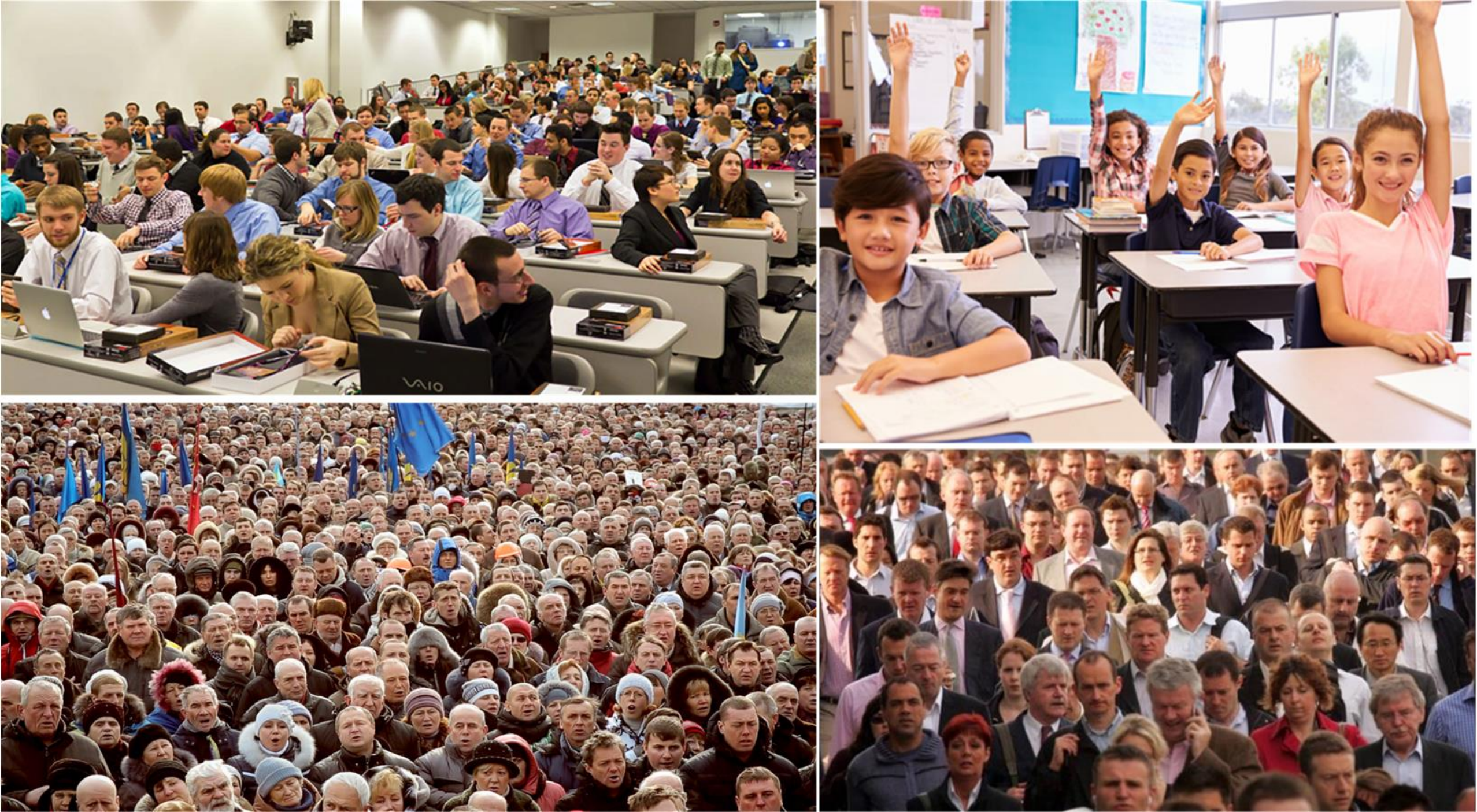}
    \caption{Faces captured with low resolutions often occur in real-world scenarios. All the pictures used in this context were sourced from \textit{www.google.com.}}
    \vspace{-10pt}
    \label{fig:SR_faces}
\end{figure}

\IEEEPARstart{M}icro-expressions, different from macro-expressions \cite{Zhang2018a}, are hidden emotions and occur over small regions of the face. The Micro-expression Recognition (\textbf{MER}) has tremendous impact on wide-range of applications in police case diagnosis and medicine fields etc~\cite{Ekman1969,  Ekm2009}. The purpose of MER is to reveal the hidden emotions of humans and help to understand people's deceitful behaviors when micro-expressions occur~\cite{Michael2010,  Ekm2009}. Recently, a variety of methods have been proposed to improve the MER performance, involving two basic principles. The first one is to classify micro-expressions through handcrafted features, \ie, geometric-based features \cite{Liong2018, Ngo2017} and appearance-based features \cite{Zhao2007,Huang2015, Huang2016, Huang2019}, while the second one is to introduce deep learning features  \cite{Xia2020, Gan2019, Zhou2019, Mao2022, Zhou2022} to learn more discriminative information from subtle micro-expressions. Although these existing MER methods have shown promising performance, they are still far from the requirement of a real-world MER system. One major reason is that existing MER methods are mostly designed and evaluated without considering the complex scenarios encountered in practice. For example,  the samples provided for MER system are recorded in laboratory conditions with high-resolution (\textbf{HR}) camera. However, in real-world scenarios, especially in crowded scenes, facial pictures suffer from the influence of low resolutions as shown in Fig. \ref{fig:SR_faces}.  In such scenario, the MER performance very sharply degrade due to the low-resolution (\textbf{LR}) samples. According to \cite{Li2019c}, it  can be explained by two following reasons. First, compared with HR samples, LR samples seriously lost the detail information of facial area.  Second,  as the LR samples are homogeneous with the high-resolution ones, LR samples cannot be directly used as the input in the testing stage. Therefore, it is vital to investigate tailor-made super-resolution algorithms for MER, especially under low-resolution scenario, promoting MER to the real-world applications.

\begin{figure*}[t!]
  \centering
  \includegraphics[width=0.9\textwidth]{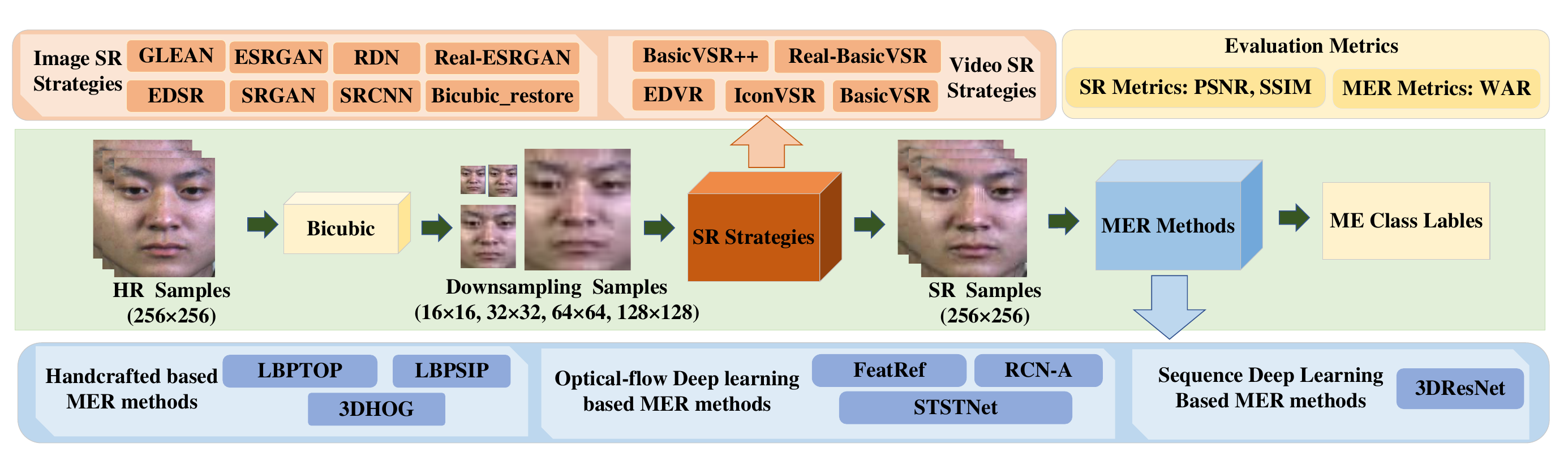}
  \caption{ The process of our benchmark. Best viewed in Color.}
  \label{fr}
\vspace{-10pt}
\end{figure*}

Prior to the research of MER, several algorithms have been proposed to deal with the SR-MER \cite{Sharma2022, Li2019c, Sharma2022a}. As far as we know, \cite{Li2019c} is the first work focused on SR-MER which used the SR method proposed in  \cite{Shi2018} to hallucinate the low-quality facial image sequence to recover the lost dynamic characteristics. Recently, Sharma \etal~\cite{Sharma2022, Sharma2022a}  transformed LR facial images into super-resolution ones by using deep learning algorithm and generative adversarial network (\textbf{GAN})~\cite{Wang2018a, Rakotonirina2020}. These methods primarily focus on performance evaluation on reconstructed samples generated by certain specific image-based super-resolution technique. Different from~\cite{Sharma2022, Li2019c, Sharma2022a}, this articl focuses on finding~\textcolor{black}{(1) the influence of LR to MER} and (2) the right and effective SR methods for MER, which provide valuable insights into the development of higher-efficiency SR-MER methods in the future. Concretely, the SOTA SR methods, \ie,  image SR methods \cite{Wang2018a, Ma2020, Chan2021, Wang2021a, Zhang2018b} and video SR methods \cite{Wang2019b, Chan2022, Chan2022a}, are considered for reconstructing ME samples. Furthermore, certain well-known feature engineered methods and deep learning based methods are investigated under LR and SR scenerios. According to the previous discussion, three questions concerned in the empirical study are raised as follows:


\begin{itemize}
\item Q1: Among those evaluated MER methods, which kind is the best for all SR strategies or a specific SR strategy across all resolutions? Through evaluating the performance of different MER methods under all SR strategies and a specific SR strategy, we aim at studying the appropriate kind of MER method if the researchers focus on the specific kind of SR strategy. It would help researchers determine which MER method is the most robust and reliable for addressing the problem caused by LR scenario. 



\item Q2: From the perspective of micro-expression samples, is video SR methods better than image SR methods?~\textcolor{black}{This involves evaluating the performance of video-based methods and image-based methods, to analyze which method is more effective for enhancing the resolution of micro-expression samples.}

\item Q3: Whether there is a proportional relationship between the quality of the super-resolved samples and the recognition results in micro-expression SR samples? \textcolor{black}{This question will be investigated by comprehensively considering the quality of the SR samples and the MER results. This involves evaluating the performance of different SR methods at different levels of quality to determine if there is a proportional relationship between the quality of the SR samples and the recognition results.}

\end{itemize}

By conducting a series of experiments, we statistically analyze these questions in-depth and provide an insight into the development of more efficient and effective SR-MER methods. To this end, in this work, we make a comprehensive survey of SOTA progress in SR, and a thorough benchmarking analysis on different SR strategies with MER methods. The experimental results on existing SR/MER works are summarized and highlighted, finally future directions are illustrated. With the summarized guidance from the survey and benchmark,
the contributions of this paper are  as follows:
\begin{itemize}
\item \textcolor{black}{We provide a comprehensive comparative evaluation of handcrafted and deep learning architectures present in existing MER literature for their performance in MER under LR scenario.}

\item \textcolor{black}{We render a study of the impact of performance variation across various SR methods, documenting their efficacy through evaluations with various low-resolutions.}

\item \textcolor{black}{We make a protocol based on synthetic low-resolution ME databases to evaluate the performance of SOTA SR and deep learning based feature extraction for MER.}

\item \textcolor{black}{We conduct a series of benchmark study based on THIRTEEN super-resolution methods and SEVEN
MER methods to investigate the SR-MER problem and
deeply discuss the experimental results. We open an interesting but challenging topic for attracting more researches and also shed a light on taking a further step towards higher-efficiency SR-MER.}

\end{itemize}

\section{BENCHMARK DETAIL}
\label{sec:benchmark_detail}
In our study, for designing the benchmark of SR-MER, we generate SR samples through those existing available SR methods, then  investigate different kinds of classifier for MER with  generated SR micro-expression samples. The detail process of our benchmark is illustrated in Fig.~\ref{fr}. Our process consists of  three major parts, including pre-processing, super-resolution samples generated, and micro-expression recognition.

\subsection{Pre-processing}
\subsubsection{Data Preparation}
SR-MER benchmark is conducted on three widely used spontaneous micro-expression databases, including SMIC-HS~\cite{Li2013},  CASME II \cite{Yan2014}, and SAMM~\cite{Davison2018}.

\textbf{SMIC-HS~\cite{Li2013}:} The SMIC-HS is recorded by a high-speed camera of 100 \textit{fps}, which has 164 micro-expression clips from 16  participants. These samples are annotated as \textit{Negative},  \textit{Positive}, and \textit{Surprise}. The sample resolution is $640 \times 480$ \textit{pixels} and  the facial area is around $190 \times 230$ \textit{pixels}.

\textbf{CASME~II~\cite{Yan2014}:} The CASME~II contains  256 samples of 7 classes (\textit{Happiness},  \textit{Surprise},  \textit{Disgust},  \textit{Sadness}, \textit{Fear}, \textit{Repression}, and \textit{Others}). All the  samples are gathered from 26 subjects. It was recorded by a camera with  200 \textit{fps}. The resolution of the samples are $640\times 480$ \textit{pixels} and the resolution of facial area is around $280 \times 340$ \textit{pixels}.

\textbf{SAMM~\cite{Davison2018}:} The SAMM database contains 159 micro-expression instances from 32 participants at 200 \textit{fps}.  The resolution of the samples are $2040 \times 1088$ \textit{pixels} and the resolution of facial area is around $400 \times 400$ \textit{pixels}. Samples in SAMM
are categorized into \textit{Happiness},  \textit{Surprise},  \textit{Disgust},  \textit{Repression},  \textit{Angry},  \textit{Fear}, \textit{Contempt}, and \textit{Others}.

Following previous MER methods~\cite{Zhou2022, Xia2020, Liong2019a}, categories in SMIC-HS are remained, while samples in CASME II and SAMM are regrouped in four classes, \ie, Positive, Negative, Surprise, and Others. The detail information of the new groups are listed in Table \ref{database}.


\begin{table*}[t!]
\caption{The sample distribution on \textit{positive}, \textit{negative}, \textit{surprise}, and \textit{others} of three databases. Subject number is given in the table.}
\small
\label{database}
\begin{center}
\setlength{\tabcolsep}{1.0mm}{
\begin{tabular}{lcccccccc}
\hline
 \multirow{2}*{Database} &\multicolumn{2}{c}{Resolution}&\multicolumn{5}{c}{Micro-Expression Category} &\multirow{2}*{Subjects}\\
 \cmidrule(r){2-3}
 \cmidrule(r){4-8}
 &Sample&Facial Area&\textit{Negative} &\textit{Positive}& \textit{Surprise} & \textit{Others}& \textbf{Total} \\
\hline
SMIC-HS~\cite{Li2013}& {$640 \times 480$} &{$190\times 230$}&70&51&43&-&164&16 \\
\hline
CASME II~\cite{Yan2014}& {$640 \times 480$} &{$280\times 340$}&{73$^{\dag}$}&{32}&{25}&{126$^{\S}$}&{256}&{26}\\
\hline
SAMM \cite{Davison2018}& {$2040 \times 1088$} &{$400\times 400$}&92$^{\sharp}$ &26&15&26&159&32\\
\hline
\multicolumn{6}{l}{\scriptsize{${\dag}$ Negative class of CASME II: Disgust, Sadness and Fear.}} \\
\multicolumn{6}{l}{\scriptsize{${\S}$ Others class of CASME II: Repression and Others. }}\\
\multicolumn{6}{l}{\scriptsize{${\sharp}$ Negative class of SAMM: Disgust, Anger, Contempt, Fear and Sadness. }}\\
\end{tabular}}
\end{center}
\vspace{-10pt}
\end{table*}

\subsubsection{Evaluation metrics}

As two major components of our benchmark are micro-expression samples reconstruction and MER with reconstructed samples, thus,  two types of evaluation metrics are considered.

 \textbf{Performance metrics for SR methods:} In our benchmark, follow with existing SR methods, the quantitative indicator Peak-Signal-to-Noise Ratio (\textbf{PSNR}) and Structural Similarity Index Measure (\textbf{SSIM}) are employed to evaluated the quality of the reconstructed micro-expression samples. It is noted that a higher PSNR value accompanies with a higher reconstructed image quality, and a higher SSIM indicates that the super-resolved image closely approaches to the ground-truth image.

 \textbf{Performance metric for MER methods:} Weight average rate score (\textbf{WAR}) is used, which are defined as follows:
\begin{equation}
\footnotesize
\text{WAR} = \frac{\sum_{k=1}^{K}TP_k}{N},
\label{war}
\end{equation}
where $K$ is the number of classes, $N$ is the total samples, and $TP_k$ is the  true positive samples in the $k$th class.

\subsubsection{Sample downscaling} Following many of the SR literature \cite{Ledig2016, Wang2018a}, to study the effectiveness of SR techniques on LR micro-expression samples, we apply Bicubic interpolation to downsample the samples from high-resolution size of $256 \times 256$ to $16 \times 16$, $32 \times 32$, $64 \times 64$, and $128 \times 128$, to obtain the LR samples.

\subsection{Super-resolution methods}
For evaluating SR methods to handle the LR micro-expression samples,  we first consider ``Bicubic\_restore"  in the SR  method for comparison with those SOTA image SR methods and video SR methods. \textcolor{black}{Specifically, ``Bicubic\_restore" uses a bicubic kernel to interpolate the LR image and produce a SR image. It is known that Bicubic interpolation is fast and easy to implement, it often results in blurry images with low-quality details. Besides ``Bicubic\_restore", other SOTA image SR and
video SR methods are introduced as following.}


\subsubsection{Image super-resolution (\textbf{ISR}) methods}

ISR is an image transformation task to reconstruct a SR image from its degraded LR one.

\begin{itemize}
\item
\textbf{SRCNN \cite{Dong2016}} was proposed by Dong~\etal~as the first CNN-based image SR model. It consists of only three layers and requires the LR image to be up-sampled using Bicubic interpolation before being processed by the network. SRCNN demonstrates that a deep CNN model is equivalent to the sparse-coding-based method, which is an example-based method for image super-resolution. 

\item
\textbf{SRGAN \cite{Ledig2016}} was proposed by combining GANs. In SRGAN, the generator $G$ is essentially a super-resolution model that is trained for producing outputs to fool the discriminator $D$. Meanwhile, $D$ is trained to distinguish between SR images and high-resolution  images. 
This process allows SRGAN to generate more realistic and visually appealing super-resolved images compared to traditional super-resolution methods.

\item
\textbf{EDSR \cite{Lim2017}} is based on ResNet. In EDSR, batch normalisation layers are removed to disable restriction of the feature values and reduce memory usage during training which allows more layers and filters to be used. Residual scaling is also used to ensure stability of the network. A multi-scale network called Multi-scale Deep Super-Resolution  is further designed. 

\item
\textbf{ESRGAN \cite{Wang2018a}} is an improvement of SRGAN. Specifically, the batch normalization layers used in SRGAN are removed, and the Residual-in-Residual Dense Block is proposed as the basic building block. 
Furthermore, inspired by the relativistic GAN approach, the adversarial loss is modified to determine the relative ``realness" of an image.  Additionally, the features computed  after the convolution layers have taken place are used to reduce the sparsity of features and better supervise brightness consistency and texture recovery.

\item
\textbf{GLEAN \cite{Chan2021}} uses pre-trained GANs as a latent bank to improve the restoration quality of large-factor image super-resolution. It directly leverages rich and diverse priors encapsulated in a pre-trained GAN. Furthermore, it only needs a single forward pass to generate the upscaled image. It can be easily incorporated in a simple encoder-bank-decoder architecture with multi-resolution skip connections. Switching the bank allows the method to deal with images from diverse categories.

\item
\textbf{RDN \cite{Zhang2018b}} aims to make full use of the hierarchical features from the original LR images by exploiting the hierarchical features from all the convolutional layers. The building module for RDN is the Residual Dense Block (RDB), which can extract abundant local features through densely connected convolutional layers. 
RDB is designed to extract and reuse features from different scales of LR images, while the RDN model leverages these features to achieve high-quality super-resolution results.

\item
\textbf{Real-ESRGAN \cite{Wang2021a}} is an extension of ESRGAN that generates synthetic images using a ``high-order" degradation process, where a degradation model is essentially applied twice. Real-ESRGAN introduces a high-order degradation modeling process to better simulate complex real-world degradation. In the synthesis process of Real-ESRGAN, common ringing and overshoot artifacts are considered. 
\end{itemize}

\subsubsection{Video super-resolution (\textbf{VSR}) methods}
VSR is known as multi-frame super-resolution. Different from ISR, it aims at addressing the issue of how to reconstruct HR images with better visual quality and finer spectral details by combining complimentary information from multiple LR counterparts.

\begin{itemize}

\item
\textbf{EDVR \cite{Wang2019b}} is a novel approach for video super-resolution. It is based on a Pyramid, Cascading, and Deformable  alignment network and a Temporal and Spatial Attention  fusion network.
It is an effective approach for video super-resolution, particularly in scenarios with large motions and diverse blur. It has shown promising results in improving the quality of super-resolved videos compared to previous state-of-the-art methods.

\item
\textbf{BasicVSR \cite{Kelvin2020}} is a video super-resolution method  that summarizes the common video super-resolution pipelines into four components: Propagation, Alignment, Aggregation, and Upsampling.
In BasicVSR, a bidirectional propagation technique is adopted for reconstruction to exploit information from the entire input video. Additionally, optical flow is used for feature warping in the alignment stage.

\item
\textbf{IconVSR \cite{Kelvin2020}} is an extension of BasicVSR, which uses BasicVSR as its backbone. IconVSR introduces the Information-Refill mechanism and Coupled Propagation components to improve the performance of BasicVSR.
The Information-Refill mechanism mitigates error accumulation during the propagation process, while Coupled Propagation facilitates information aggregation by jointly propagating the information from multiple frames. 

\item
\textbf{BasicVSR++ \cite{Chan2022}} is an advanced version of BasicVSR that improves upon the original method by introducing second-order grid propagation and flow-guided deformable alignment. Both components enable better propagation and aggregation of information in the video super-resolution process. Therefore, BasicVSR++ is able to exploit  spatiotemporal information in video super-resolution more effectively, resulting in higher-quality super-resolved videos.
\item
\textbf{RealBasicVSR \cite{Chan2022a}} is an extension work of BasicVSR from non-blind VSR to real-world VSR,  attempting to balance the tradeoff between detail synthesis and artifact suppression. This work focuses at the two aspects in real-world VSR: (i) proposing a stochastic degradation scheme for the issue of speed-performance equilibrium; (ii) suggesting that networks trained with longer sequences rather than larger batches to handle the batch-length balance issue. Also, in RealBasicVSR, Chan \etal\ found an image pre-cleaning stage indispensable to reduce noises and artifacts prior to propagation.
\end{itemize}

\subsection{Micro-expression recognition methods}

 We up-scale the LR samples into HR samples, followed by MER method. MER methods are introduced in our benchmark to classify the restorated micro-expression samples into corresponding categories. several well-used MER methods are employed in our benchmark.

\subsubsection{Handcrafted based MER methods}
Three handcrafted features are considered in the study.
\begin{itemize}
\item
\textbf{LBPTOP \cite{Zhao2007}} is the most widely used appearance-based feature for MER. It combines temporal features and  spatial features from three orthogonal planes of the image sequence. 

\item
\textbf{LBPSIP \cite{Wang2014}} is a spatiotemporal descriptor utilizing six intersection points to suppress redundant information in LBPTOP and preserve more efficient computational complexity.

\item
\textbf{3DHOG \citep{Polikovsky2009}} is an appearance-based feature that is obtained by counting occurrences of gradient orientation in localized portions of a given image sequence. In 3DHOG, face is devided into 12 regions
selected through manual annotation of points on the face
and then a rectangle was centred on these points. 3DHOG was used to recognise motion in each region.

\end{itemize}
\subsubsection{Optical-flow deep learning based MER methods}

Recently, with the advancement of deep learning, shallow networks with motion features obtained from the onset and apex frames can efficiently reduce over-fitting and improve the recognition performance in MER.

\begin{itemize}
\item
\textbf{Recurrent convolutional network with attention unit (RCNA) \cite{Xia2020a}} is a type of RCN proposed by Xia \etal. With the integration of RCN without increasing any learnable parameters, RCNA can enhance the representation ability in various perspectives.

\item
\textbf{Shallow Triple Stream Three-dimensional network (STSTNet) \cite{Liong2019a}} is a two-layer neural network that is capable of learning the features from three optical flow features, \ie, optical strain, horizontal and vertical optical flow fields computed based on the onset and apex frames of each
video.

\item
\textbf{Feature Refinement (FeatRef) \cite{Zhou2022}} is proposed based on a shallow network named Dual-Inception Network \cite{Zhou2019}. With expression-specific feature learning and fusion, FeatRef can successfully obtains salient and discriminative features for MER without any data augmentation.

\end{itemize}
\subsubsection{Sequence deep learning based MER methods}
Besides those methods proposed specific for MER, we also employ a generic sequence-fed method 3DResNet to evaluate the performance of low-resolution MER with SR samples.
\begin{itemize}
\item
\textbf{3DResNet \cite{Hara2017}} is proposed based on ResNet. It is initially used for action recognition. The difference between 3DResNet and original ResNet \cite{He2016a} is the number of dimensions of convolutional kernels and pooling. 3D ResNet perform 3D convolution and 3D pooling.
With the input of sequence, 3DResNet can learning spatio-temporal features in an end-to-end way.
\end{itemize}

\begin{table*}[t!]
\small
  \centering
  \caption{Performance evaluation of different SR methods for SR-MER on the CASME II database. The best result under specific SR method is in underline. The best average result among MER methods is in bold. The best average result of SR method over $m\times m$ and $m$to$256$ is in \textcolor{red}{red color}, where $m$ is the image resolution. Additionally, the average result of all MER methods at the resolution of $256\times256$ is in \textcolor{blue}{blue color}.}

    \begin{tabular}{|cc|rrrrrrc|r|}
    \hline
    \multicolumn{1}{|c|}{\multirow{3}[4]{*}{ SR  Method}} & \multirow{2}[4]{*}{Resolution} & \multicolumn{8}{c|}{MER Methods} \\
\cline{3-10}    \multicolumn{1}{|c|}{} &       & \multicolumn{1}{c}{LBPTOP} & \multicolumn{1}{c}{LBPSIP} & \multicolumn{1}{c}{3DHOG} & \multicolumn{1}{c}{RCNA} & \multicolumn{1}{c}{STSTNet} & \multicolumn{1}{c}{FeatRef} & \multicolumn{1}{c}{3DResNet} & \multicolumn{1}{c|}{Avg.} \\
    \cline{2-10}
    \multicolumn{1}{|c|}{} & 256$\times$256 & 0.471 & 0.482 & 0.373 & 0.626 & 0.540 & \underline{0.629} & 0.408 & \textcolor{blue}{0.504} \\
    \hline
    \multicolumn{1}{|c|}{Bicubic} & 16$\times$16 & \underline{0.435}& 0.357 & 0.271 & 0.365 & 0.411 & 0.397 & 0.419 & 0.379 \\
    \multicolumn{1}{|c|}{Bicubic\_restore} & 16to256 & 0.396 & 0.420 & 0.333 & \underline{0.461} & 0.445 & 0.430 & 0.416 & 0.414 \\
    \multicolumn{1}{|c|}{EDSR} & 16to256 & \underline{0.447} & \underline{0.447} & 0.380 & 0.373 & 0.432 & 0.406 & 0.408 & 0.413 \\
    \multicolumn{1}{|c|}{RDN} & 16to256 & 0.431 & 0.431 & 0.396 & 0.396 & \underline{0.450} & 0.412 & 0.385 & 0.415 \\
    \multicolumn{1}{|c|}{GLEAN} & 16to256 & 0.494 & \underline{0.494} & 0.298 & 0.400 & 0.463 & 0.414 & 0.382 & \textcolor{red}{0.421} \\
    \hline
    \multicolumn{1}{|c|}{Bicubic} & 32$\times$32 & 0.369 & 0.420 & 0.333 & 0.473 & 0.468 & \underline{0.475} & 0.373 & 0.416 \\
    \multicolumn{1}{|c|}{Bicubic\_restore} & 32to256 & 0.400 & 0.443 & 0.337 & \underline{0.547} & 0.501 & 0.541 & 0.406 & 0.454 \\
    \multicolumn{1}{|c|}{EDSR} & 32to256 & 0.416 & 0.451 & 0.337 & \underline{0.513} & 0.479 & 0.488 & 0.433 & 0.445 \\
    \multicolumn{1}{|c|}{RDN} & 32to256 & 0.408 & 0.471 & 0.314 & \underline{0.533} & 0.484 & 0.489 & 0.423 & 0.446 \\
    \multicolumn{1}{|c|}{GLEAN} & 32to256 & 0.412 & 0.435 & 0.345 & \underline{0.560} & 0.508 & 0.539 & 0.444 & \textcolor{red}{0.463} \\
    \hline
    \multicolumn{1}{|c|}{Bicubic} & 64$\times$64 & 0.396 & 0.467 & 0.365 & 0.586 & 0.554 & \underline{0.588} & 0.411 & 0.481 \\
    \multicolumn{1}{|c|}{Bicubic\_restore} & 64to256 & 0.435 & 0.471 & 0.427 & 0.580 & 0.519 & \underline{0.583} & 0.413 & 0.490 \\
    \multicolumn{1}{|c|}{EDSR} & 64to256 & 0.451 & 0.478 & 0.388 & 0.570 & 0.525 & \underline{0.583} & 0.412 & 0.487 \\
    \multicolumn{1}{|c|}{RDN} & 64to256 & 0.451 & 0.478 & 0.365 & 0.582 & 0.522 & \underline{0.586} & 0.426 & 0.487 \\
    \multicolumn{1}{|c|}{GLEAN} & 64to256 & 0.494 & 0.494 & 0.337 & \underline{0.562} & 0.507 & 0.552 & 0.387 & 0.476 \\
    \multicolumn{1}{|c|}{SRCNN} & 64to256 & 0.443 & 0.486 & 0.455 & 0.567 & 0.523 & \underline{0.603} & 0.416 & 0.499 \\
    \multicolumn{1}{|c|}{SRGAN} & 64to256 & 0.451 & 0.482 & 0.369 & 0.578 & 0.509 & \underline{0.590} & 0.414 & 0.485 \\
    \multicolumn{1}{|c|}{ESRGAN} & 64to256 & 0.490 & 0.482 & 0.353 & \underline{0.565} & 0.503 & 0.554 & 0.451 & 0.486 \\
    \multicolumn{1}{|c|}{Real-ESRGAN} & 64to256 & 0.451 & 0.475 & 0.388 & 0.563 & 0.490 & \underline{0.578} & 0.413 & 0.480 \\
    \multicolumn{1}{|c|}{EDVR} & 64to256 & 0.431 & 0.478 & 0.388 & \underline{0.567} & 0.510 & 0.560 & 0.430 & 0.481 \\
    \multicolumn{1}{|c|}{BasicVSR} & 64to256 & 0.467 & 0.482 & 0.416 & \underline{0.548} & 0.493 & 0.545 & 0.425 & 0.482 \\
    \multicolumn{1}{|c|}{IconVSR} & 64to256 & 0.451 & 0.482 & 0.420 & \underline{0.587} & 0.525 & 0.571 & 0.416 & 0.493 \\
    \multicolumn{1}{|c|}{BasicVSR++} & 64to256 & 0.459 & 0.478 & 0.388 & 0.593 & 0.536 & \underline{0.600} & 0.450 & \textcolor{red}{0.501} \\
    \multicolumn{1}{|c|}{RealBasicVSR} & 64to256 & 0.494 & 0.494 & 0.376 & \underline{0.514} & 0.499 & 0.508 & 0.366 & 0.465 \\
   \hline
    \multicolumn{1}{|c|}{Bicubic} & 128$\times$128 & 0.459 & 0.478 & 0.341 & \underline{0.632} & 0.531 & 0.627 & 0.404 & \textcolor{red}{0.496} \\
    \multicolumn{1}{|c|}{Bicubic\_restore} & 128to256 & 0.478 & 0.482 & 0.353 & \underline{0.646} & 0.513 & 0.618 & 0.371 & 0.495 \\
    \multicolumn{1}{|c|}{EDSR} & 128to256 & 0.482 & 0.490 & 0.345 & \underline{0.619} & 0.514 & 0.616 & 0.398 & 0.495 \\
    \multicolumn{1}{|c|}{RDN} & 128to256 & 0.482 & 0.486 & 0.349 & 0.615 & 0.514 & \underline{0.624} & 0.378 & 0.493 \\
    \multicolumn{1}{|c|}{GLEAN} & 128to256 & 0.490 & 0.494 & 0.412 & \underline{0.569} & 0.518 & 0.561 & 0.398 & 0.492 \\
    \hline
    \multicolumn{2}{|c|}{\textbf{Avg.}} & 0.448 & 0.467 & 0.365 & \textbf{0.543} & 0.499 & 0.542 & 0.409 &  \\
    \hline
    \end{tabular}%
    \vspace{-10pt}
  \label{tab:war_casme2}%
\end{table*}%

\begin{table*}[t!]
  \centering
  \caption{Performance evaluation of MER methods with different SR strategies on the SAMM database. The best result under specific SR method is in underline. The best average result among MER methods is in bold. The best average result of SR method over $m\times m$ and $m$to$256$ is in \textcolor{red}{red color}, where $m$ is the image resolution. Additionally, the average result of all MER methods at the resolution of $256\times256$ is in \textcolor{blue}{blue color}.}
  \small
    \begin{tabular}{|cc|rrrrrrr|r|}
    \hline
    \multicolumn{1}{|c|}{\multirow{3}[4]{*}{ SR  Method}} & \multirow{2}[4]{*}{Resolution} & \multicolumn{8}{c|}{MER Methods} \\
\cline{3-10}    \multicolumn{1}{|c|}{} &       & \multicolumn{1}{c}{LBPTOP} & \multicolumn{1}{c}{LBPSIP} & \multicolumn{1}{c}{3DHOG} & \multicolumn{1}{c}{RCNA} & \multicolumn{1}{c}{STSTNet} & \multicolumn{1}{c}{FeatRef} & \multicolumn{1}{c|}{3DResNet} & \multicolumn{1}{c|}{Avg.} \\
    \cline{2-10}
    \multicolumn{1}{|c|}{} & 256$\times$256 & 0.572 & 0.579 & 0.509 & \underline{0.596} & 0.560 & 0.589 & 0.512 & \textcolor{blue}{0.560} \\
    \hline
    \multicolumn{1}{|c|}{Bicubic} & 16$\times$16 & 0.440 & 0.497 & 0.528 & 0.419 & \underline{0.529} & 0.444 & 0.511 & 0.481 \\
    \multicolumn{1}{|c|}{Bicubic\_restore} & 16to256 & 0.547 & 0.560 & 0.478 & 0.477 & \underline{0.549} & 0.504 & 0.449 & 0.509 \\
    \multicolumn{1}{|c|}{EDSR} & 16to256 & \underline{0.553} & 0.566 & 0.453 & \underline{0.572} & 0.542 & 0.473 & 0.444 & \textcolor{red}{0.515} \\
    \multicolumn{1}{|c|}{RDN} & 16to256 & 0.553 & \underline{0.572} & 0.428 & 0.422 & 0.533 & 0.462 & 0.454 & 0.489 \\
    \multicolumn{1}{|c|}{GLEAN} & 16to256 & 0.572 & \underline{0.579} & 0.453 & 0.428 & 0.554 & 0.510 & 0.436 & 0.504 \\
    \hline
    \multicolumn{1}{|c|}{Bicubic} & 32$\times$32 & 0.528 & \underline{0.579} & 0.409 & 0.453 & 0.538 & 0.438 & 0.506 & 0.493 \\
    \multicolumn{1}{|c|}{Bicubic\_restore} & 32to256 & \underline{0.591} & \underline{0.591} & 0.358 & 0.478 & 0.538 & 0.511 & 0.506 & 0.511 \\
    \multicolumn{1}{|c|}{EDSR} & 32to256 & \underline{0.597} & 0.585 & 0.453 & 0.452 & 0.562 & 0.477 & 0.463 & 0.513 \\
    \multicolumn{1}{|c|}{RDN} & 32to256 & \underline{0.591} & 0.585 & 0.472 & 0.463 & 0.549 & 0.480 & 0.471 & \textcolor{red}{0.516} \\
    \multicolumn{1}{|c|}{GLEAN} & 32to256 & \underline{0.604} & 0.585 & 0.434 & 0.456 & 0.553 & 0.509 & 0.460 & 0.514 \\
    \hline
    \multicolumn{1}{|c|}{Bicubic} & 64$\times$64 & 0.560 & \underline{0.579} & 0.465 & 0.497 & 0.538 & 0.521 & 0.453 & 0.516 \\
    \multicolumn{1}{|c|}{Bicubic\_restore} & 64to256 & \underline{0.591} & 0.572 & 0.403 & 0.524 & 0.551 & 0.526 & 0.494 & 0.523 \\
    \multicolumn{1}{|c|}{EDSR} & 64to256 & \underline{0.572} & \underline{0.572} & 0.403 & 0.471 & 0.552 & 0.504 & 0.438 & 0.502 \\
    \multicolumn{1}{|c|}{RDN} & 64to256 & \underline{0.572} & \underline{0.572} & 0.403 & 0.513 & 0.555 & 0.518 & 0.445 & 0.511 \\
    \multicolumn{1}{|c|}{GLEAN} & 64to256 & 0.572 & \underline{0.579} & 0.440 & 0.515 & 0.553 & 0.516 & 0.462 & 0.520 \\
    \multicolumn{1}{|c|}{SRCNN} & 64to256 & \underline{0.572} & \underline{0.572} & 0.409 & 0.508 & 0.549 & 0.509 & 0.483 & 0.515 \\
    \multicolumn{1}{|c|}{SRGAN} & 64to256 & \underline{0.572} & \underline{0.572} & 0.403 & 0.522 & 0.555 & 0.524 & 0.445 & 0.513 \\
    \multicolumn{1}{|c|}{ESRGAN} & 64to256 & \underline{0.572} & \underline{0.572} & 0.390 & 0.519 & 0.549 & 0.525 & 0.404 & 0.504 \\
    \multicolumn{1}{|c|}{Real-ESRGAN} & 64to256 & \underline{0.585} & 0.572 & 0.434 & 0.544 & 0.561 & 0.551 & 0.479 & \textcolor{red}{0.532} \\
    \multicolumn{1}{|c|}{EDVR} & 64to256 & \underline{0.572} & \underline{0.572} & 0.415 & 0.519 & 0.540 & 0.514 & 0.425 & 0.508 \\
    \multicolumn{1}{|c|}{BasicVSR} & 64to256 & 0.566 & \underline{0.572} & 0.421 & 0.528 & 0.558 & 0.497 & 0.475 & 0.517 \\
    \multicolumn{1}{|c|}{IconVSR} & 64to256 & \underline{0.553} & \underline{0.553} & 0.333 & 0.500 & 0.530 & 0.500 & 0.469 & 0.491 \\
    \multicolumn{1}{|c|}{BasicVSR++} & 64to256 & \underline{0.572} & \underline{0.572} & 0.409 & 0.509 & 0.552 & 0.503 & 0.443 & 0.508 \\
    \multicolumn{1}{|c|}{RealBasicVSR} & 64to256 & 0.572 & \underline{0.579} & 0.459 & 0.466 & 0.558 & 0.501 & 0.478 & 0.516 \\
    \hline
    \multicolumn{1}{|c|}{Bicubic} & 128$\times$128 & \underline{0.572} & \underline{0.572}& 0.478 & 0.525 & 0.550 & 0.522 & 0.484 & 0.529 \\
    \multicolumn{1}{|c|}{Bicubic\_restore} & 128to256 & \underline{0.572} & \underline{0.572} & 0.516 & 0.561 & 0.552 & 0.565 & 0.480 & 0.546 \\
    \multicolumn{1}{|c|}{EDSR} & 128to256 & \underline{0.572} & \underline{0.572} & 0.528 & 0.520 & 0.553 & 0.562 & 0.491 & 0.543 \\
    \multicolumn{1}{|c|}{RDN} & 128to256 & \underline{0.572} & \underline{0.572} & 0.503 & 0.569 & 0.566 & 0.557 & 0.492 & \textcolor{red}{0.547} \\
    \multicolumn{1}{|c|}{GLEAN} & 128to256 & 0.572 & \underline{0.579} & 0.434 & 0.551 & 0.561 & 0.522 & 0.491 & 0.530 \\
    \hline
    \multicolumn{2}{|c|}{Avg.} & 0.567 & {\textbf{0.572}} & 0.441 & 0.503 & 0.550 & 0.511 & 0.468 &  \\
    \hline
    \end{tabular}%
   \vspace{-10pt}
  \label{tab:war_samm}%
\end{table*}%

\begin{table*}[htbp]
  \centering
  \caption{Performance evaluation of MER methods with different SR strategies on the SMIC-HS database. The best result under specific SR method is in underline. The best average result among MER methods is in bold. The best average result of SR method over $m\times m$ and $m$to$256$ is in \textcolor{red}{red color}, where $m$ is the image resolution. Additionally, the average result of all MER methods at the resolution of $256\times256$ is in \textcolor{blue}{blue color}.}
  \small
    \begin{tabular}{|cc|rrrrrrr|r|}
    \hline
    \multicolumn{1}{|c|}{\multirow{3}[4]{*}{ SR  Method}} & \multirow{2}[4]{*}{Resolution} & \multicolumn{8}{c|}{MER Methods} \\
\cline{3-10}    \multicolumn{1}{|c|}{} &       & \multicolumn{1}{c}{LBPTOP} & \multicolumn{1}{c}{LBPSIP} & \multicolumn{1}{c}{3DHOG} & \multicolumn{1}{c}{RCNA} & \multicolumn{1}{c}{STSTNet} & \multicolumn{1}{c}{FeatRef} & \multicolumn{1}{c|}{3DResNet} & \multicolumn{1}{l|}{Avg.} \\
    \cline{2-10}
    \multicolumn{1}{|c|}{} & 256$\times$256 & 0.372 & 0.317 & 0.390 & 0.565 & 0.522 & \underline{0.549} & 0.424 & \multicolumn{1}{c|}{\textcolor{blue}{0.449}} \\
    \hline
    \multicolumn{1}{|c|}{Bicubic} & 16 $\times$16 & 0.323 & 0.317 & 0.244 & 0.375 & 0.384 & \underline{0.391} & 0.301 & \multicolumn{1}{c|}{0.334} \\
    \multicolumn{1}{|c|}{Bicubic\_restore} & 16to256 & 0.348 & 0.305 & \underline{0.409} & 0.387 & 0.373 & 0.393 & 0.318 & \multicolumn{1}{c|}{0.362} \\
    \multicolumn{1}{|c|}{EDSR} & 16to256 & 0.372 & 0.280 & 0.366 & 0.423 & \underline{0.436} & 0.401 & 0.373 & \multicolumn{1}{c|}{0.379} \\
    \multicolumn{1}{|c|}{RDN} & 16to256 & \underline{0.366} & 0.287 & 0.317 & 0.354 & 0.313 & 0.351 & 0.285 & \multicolumn{1}{c|}{0.325} \\
    \multicolumn{1}{|c|}{GLEAN} & 16to256 & 0.384 & \underline{0.427} & 0.341 & 0.399 & 0.369 & 0.384 & 0.401 & \multicolumn{1}{c|}{\textcolor{red}{0.386}} \\
    \hline
    \multicolumn{1}{|c|}{Bicubic} & 32$\times$32 & 0.268 & 0.390 & 0.244 & \underline{0.460} & 0.414 & 0.460 & 0.310 & \multicolumn{1}{c|}{0.364} \\
    \multicolumn{1}{|c|}{Bicubic\_restore} & 32to256 & 0.299 & 0.299 & 0.287 & \underline{0.445} & 0.439 & 0.443 & 0.313 & \multicolumn{1}{c|}{0.361} \\
    \multicolumn{1}{|c|}{EDSR} & 32to256 & 0.293 & 0.335 & 0.360 & 0.443 & \underline{0.451} & 0.418 & 0.305 & \multicolumn{1}{c|}{0.372} \\
    \multicolumn{1}{|c|}{RDN} & 32to256 & 0.305 & 0.335 & 0.305 & 0.437 & \underline{0.450} & 0.420 & 0.320 & \multicolumn{1}{c|}{0.367} \\
    \multicolumn{1}{|c|}{GLEAN} & 32to256 & 0.384 & 0.329 & 0.323 & \underline{0.469} & 0.430 & 0.450 & 0.313 & \multicolumn{1}{c|}{\textcolor{red}{0.386}} \\
    \hline
    \multicolumn{1}{|c|}{Bicubic} & 64$\times$64 & 0.335 & 0.280 & 0.238 & 0.488 & 0.431 & \underline{0.504} & 0.473 & \multicolumn{1}{c|}{0.393} \\
    \multicolumn{1}{|c|}{Bicubic\_restore} & 64to256 & 0.360 & 0.360 & 0.335 & \underline{0.536} & 0.457 & 0.485 & 0.355 & \multicolumn{1}{c|}{0.413} \\
    \multicolumn{1}{|c|}{EDSR} & 64to256 & 0.372 & 0.348 & 0.305 & \underline{0.505} & 0.473 & 0.495 & 0.434 & \multicolumn{1}{c|}{0.419} \\
    \multicolumn{1}{|c|}{RDN} & 64to256 & 0.396 & 0.348 & 0.329 & \underline{0.530} & 0.461 & 0.491 & 0.401 & \multicolumn{1}{c|}{\textcolor{red}{0.422}} \\
    \multicolumn{1}{|c|}{GLEAN} & 64to256 & 0.323 & 0.335 & 0.299 & \underline{0.493} & 0.420 & 0.465 & 0.478 & \multicolumn{1}{c|}{0.402} \\
    \multicolumn{1}{|c|}{SRCNN} & 64to256 & 0.384 & 0.348 & 0.305 & \underline{0.531} & 0.464 & 0.502 & 0.449 & \multicolumn{1}{c|}{0.426} \\
    \multicolumn{1}{|c|}{SRGAN} & 64to256 & 0.390 & 0.335 & 0.287 & \underline{0.517} & 0.473 & 0.475 & 0.393 & \multicolumn{1}{c|}{0.410} \\
    \multicolumn{1}{|c|}{ESRGAN} & 64to256 & 0.372 & 0.317 & 0.335 & \underline{0.455} & 0.420 & 0.431 & 0.428 & \multicolumn{1}{c|}{0.394} \\
    \multicolumn{1}{|c|}{Real-ESRGAN} & 64to256 & 0.384 & 0.372 & 0.293 & \underline{0.522} & 0.441 & 0.495 & 0.373 & \multicolumn{1}{c|}{0.411} \\
    \multicolumn{1}{|c|}{EDVR} & 64to256 & 0.372 & 0.341 & 0.323 & \underline{0.508} & 0.445 & 0.489 & 0.371 & \multicolumn{1}{c|}{0.407} \\
    \multicolumn{1}{|c|}{BasicVSR} & 64to256 & 0.360 & 0.299 & 0.317 & \underline{0.449} & 0.405 & 0.410 & 0.402 & \multicolumn{1}{c|}{0.378} \\
    \multicolumn{1}{|c|}{IconVSR} & 64to256 & 0.348 & 0.329 & 0.299 & \underline{0.483} & 0.455 & \underline{0.483} & 0.380 & \multicolumn{1}{c|}{0.397} \\
    \multicolumn{1}{|c|}{BasicVSR++} & 64to256 & 0.378 & 0.317 & 0.360 & \underline{0.486} & 0.468 & 0.471 & 0.383 & \multicolumn{1}{c|}{0.409} \\
    \multicolumn{1}{|c|}{RealBasicVSR} & 64to256 & 0.396 & 0.384 & 0.287 & 0.488 & 0.443 & \underline{0.505} & 0.385 & \multicolumn{1}{c|}{0.413} \\
    \midrule
    \multicolumn{1}{|c|}{Bicubic} & 128$\times$128 & 0.360 & 0.305 & 0.415 & \underline{0.543} & 0.498 & 0.523 & 0.522 & \multicolumn{1}{c|}{0.452} \\
    \multicolumn{1}{|c|}{Bicubic\_restore} & 128to256 & 0.378 & 0.335 & 0.366 & \underline{0.537} & 0.519 & 0.531 & 0.527 & \multicolumn{1}{c|}{0.456} \\
    \multicolumn{1}{|c|}{EDSR} & 128to256 & 0.384 & 0.341 & 0.372 & \underline{0.556} & 0.523 & \underline{0.556} & 0.507 & \multicolumn{1}{c|}{\textcolor{red}{0.463}} \\
    \multicolumn{1}{|c|}{RDN} & 128to256 & 0.372 & 0.354 & 0.341 & \underline{0.526} & 0.509 & 0.522 & 0.509 & \multicolumn{1}{c|}{0.448} \\
    \multicolumn{1}{|c|}{GLEAN} & 128to256 & 0.335 & 0.311 & 0.317 & 0.521 & 0.490 & \underline{0.523} & 0.409 & \multicolumn{1}{c|}{0.415} \\
    \hline
    \multicolumn{2}{|c|}{Avg.} & 0.357 & 0.333 & 0.324 & \textbf{0.481} & 0.446 & 0.467 & 0.395 & \multicolumn{1}{c|}{} \\
    \hline
    \end{tabular}%
  \label{tab:war_smichs}%
\end{table*}%

\section{Experiments and Discussion}
\subsection{Implementation Details}

We conduct benchmark SR-MER experiments under the process described in Section  \ref{sec:benchmark_detail}. To investigate the influence of LR to MER, firstly, we set $256 \times 256$ pixel for all ME samples. Then we downsample $256 \times 256$-resolution samples to four resolution cases including $16 \times 16$, $32 \times 32$, $64 \times 64$, and  $128 \times 128$. On the other hand, to analyze the influence of SR in our study, we super-resolve those LR samples to  $256 \times 256$ by the SOTA SR algorithms. Subsequently, we feed those SR samples into MER methods to analyze performance of MER.

For all the SR strategies, we follow the settings in the original literature. For MER methods,  all the experiments are evaluated under the Leave-One-Subject-Out (\textbf{LOSO}) protocol. The parameters are set as following,

1) For LBPTOP \cite{Zhao2007}, the neighboring radii $R$ for the XY, XT and YT planes are set to 1, 1 and 4, respectively.  The number of neighboring points $P$ is set to 4 for all planes, and 5 $\times$ 5 nonoverlapping blocks are used. For LBPSIP \cite{Wang2014}, the neighboring radius $R$ is set to 1 and 3. For 3DHOG, the number of divisions T domains is set as 1. To offer a fair comparison, we use linear kernel based SVM throughout the evaluation experiments.

2) We extract TV-L1 optical flow from the onset and apex frames, and then feed the optical flow images into RCNA, STSTNet, and FeatRef. For RCNA,  the momentum is set to 0.9 and weight decay 0.0005 in stochastic gradient decent (SGD) with
momentum. The learning rate is set to $10^{-4}$. The stopping
criterion for SGD loss is set to 0.5 for iterations and maximum
iteration number is set to 500. For STSTNet, we set the learning rate as $5 \times 10^{-5}$, and the maximum number of epochs set to 500. For FeatRef, the learning rate and loss function weight factor $\lambda$ are set as $10^{-3}$ and 0.85, respectively.

3) For 3DResNet \cite{Hara2017}, as the input of model is sequence, we first use TIM to normalize the number of frames in each sample as 16. The learning rate is set as $10^{-4}$.

\begin{figure*}[t!]
    \centering
    \includegraphics[width=0.98\textwidth]{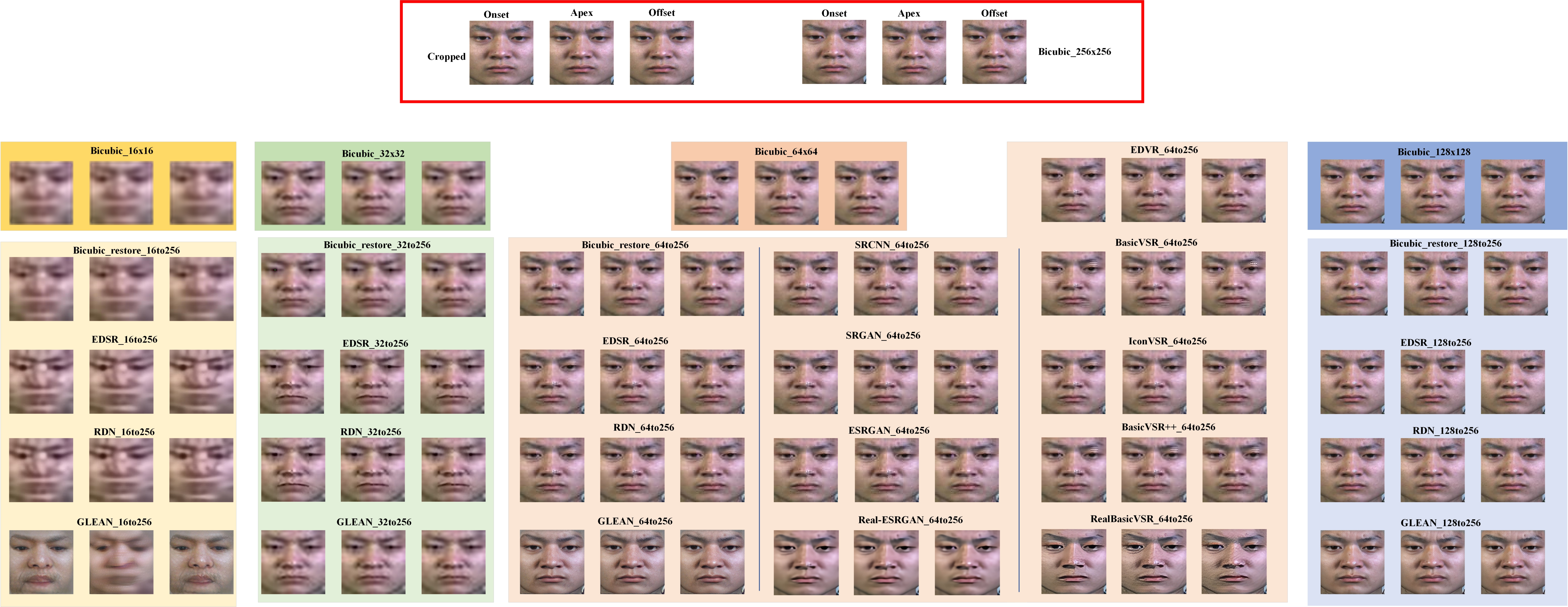}
    \caption{Comparison of reconstruction results at different
resolutions using different SR methods. Better viewed in zoom.}
    \label{fig:LR_SR_example}
\end{figure*}

\begin{figure*}[b]
     \centering
     \subfigure[]{
             \includegraphics[width=0.48\textwidth]{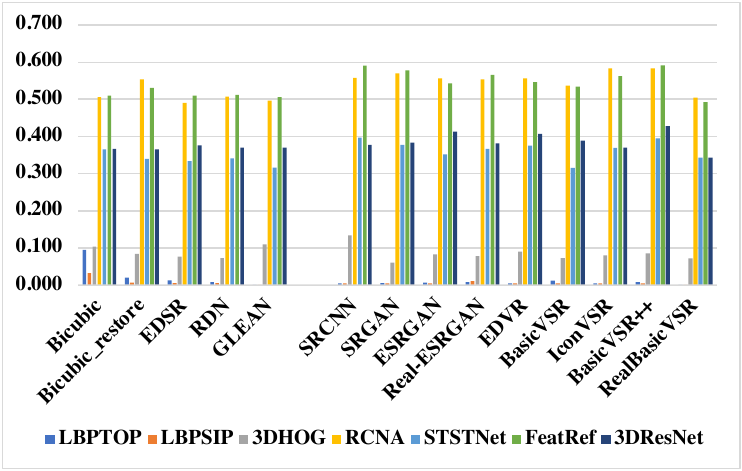}
         \label{fig:avg_war_casme2}}
     \subfigure[]{
    \includegraphics[width=0.48\textwidth]{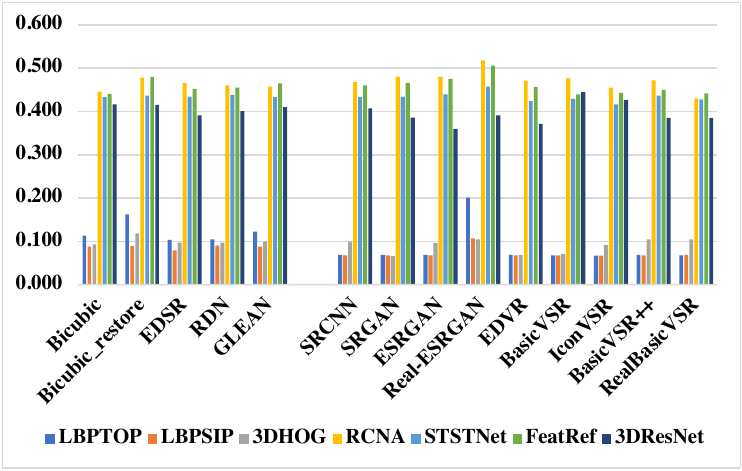}
         \label{fig:avg_war_samm}
     }
     \subfigure[]{
         \includegraphics[width=0.48\textwidth]{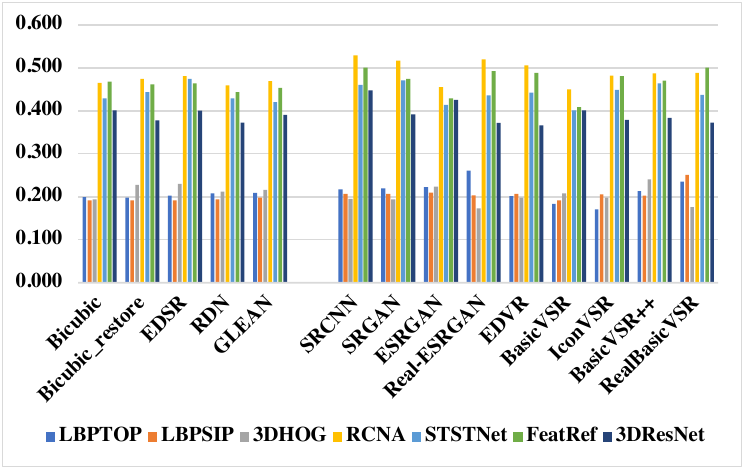}
         \label{fig:avg_war_smichs}
     }
        \caption{Average performance of each specific MER method under each SR method among all resolutions on (a) CASME II, (b) SAMM, and (c) SMIC-HS databases.}
        \label{fig:avg_war}
\end{figure*}

\subsection{\textcolor{black}{Effect of Low-resolution}}
\label{subsec:low-res_effect}

\textcolor{black}{To analyze the impact of LR on MER performance more clearly, we compare the average performance of all MER methods in Tables~\ref{tab:war_casme2},~\ref{tab:war_samm}, and~\ref{tab:war_smichs} under the resolution of $256 \times 256$ and four LR cases (\ie, we use Bicubic interpolation to downscale resolution of  $256 \times 256$ to resolutions of $16 \times 16$, $32 \times 32$, $64 \times 64$, and $128 \times 128$). The comparative results will provide information about the specific degree of the impact of LR on MER performance, thereby helping us better understand MER performance under different resolutions. }

\textcolor{black}{Firstly, according to the last column of Table~\ref{tab:war_casme2},~\ref{tab:war_samm}, and~\ref{tab:war_smichs}, except the case under the resolution of $128 \times 128$ in SMIC-HS database, the performance in most of LR cases on three micro-expression databases are significantly lower than those under the resolution of $256 \times 256$. For example, in CASME II database, the average performance under the resolution of $256 \times 256$ is 0.504, while those under the resolutions of $16 \times 16$, $32 \times 32$, $64 \times 64$, and $128 \times 128$ are 0.379, 0.416, 0.481, and 0.496, respectively. The analysis on SAMM database is the same to that on CASME II database. The performance comparison demonstrates that the lower the resolution, the worse the performance.} 

\textcolor{black}{In summary, different resolutions have a significant impact on the performance of MER. Extremely low resolutions can indeed cause a significant decrease in recognition performance. This conclusion will be validated in Section \ref{subsubsec:specific strategy} with the three-factors analysis. Therefore, it is necessary to investigate  the SR MER issue.}

\subsection{Experiment Analysis of SR methods}
\label{subsec:Analysis of SR methods}

\textcolor{black}{Due to the VSR methods can only handle $64\text{to}256$ case, we compare the VSR methods and ISR methods on $64\times 64$ and $64\text{to}256$ in Table \ref{tab:war_casme2},~\ref{tab:war_samm}, and~\ref{tab:war_smichs}.} Among VSR methods,~\textcolor{black}{only BasicVSR++ achieved better performance than those ISR methods in CASME II database}. \textcolor{black}{In order to further analyze both kinds of SR methods, we calculate the average performance along seven ISR methods / five VSR methods. The average results are 0.486 and 0.484 on CASME II database, 0.514 and 0.508 on SAMM database, 0.412 and 0.401 on SMIC-HS database for ISR methods and  VSR methods, respectively. It is seen that the image-based category gains the better performance than the video-based one. This may be explained by that ISR methods can generate higher resolution images, thus providing more micro-expression detail information. Although videos can provide rich motion information, subtle facial micro-expression may be blurry or missing in LR videos, which can to some extent affect recognition performance.} These experimental results indicate that ISR methods is more suitable for MER than VSR methods.



\begin{table*}[t!]
  \centering
  \small
  \caption{Performance comparison of SR methods in terms of PSNR and SSIM on three micro-expression databases. The best results on each database and a specific $m\text{to}256$ resolution are in bold. The second ranked one is in underline.}
    \begin{tabular}{|c|c|rr|rr|rr|}
    \hline
    \multirow{2}[4]{*}{ SR  Method} & \multirow{2}[4]{*}{Resolution} & \multicolumn{2}{c|}{CASME II} & \multicolumn{2}{c|}{SAMM} & \multicolumn{2}{c|}{SMIC} \\
\cline{3-8}          &       & \multicolumn{1}{c}{PSNR} & \multicolumn{1}{c|}{SSIM} & \multicolumn{1}{c}{PSNR} & \multicolumn{1}{c|}{SSIM} & \multicolumn{1}{c}{PSNR} & \multicolumn{1}{c|}{SSIM} \\
    \hline
    Bicubic\_restore & 16to256 & \underline{23.32} & \textbf{0.7584} & 22.66 & \textbf{0.7004} & 23.46 & \textbf{0.7597} \\
    EDSR  & 16to256 & \textbf{23.93} & \underline{0.7453} & \underline{23.00} & 0.6968 & \textbf{23.72} & \underline{0.7560} \\
    RDN   & 16to256 & \textbf{23.93} & 0.7450 & \textbf{23.02} & \underline{0.6983} & \underline{23.67} & 0.7545 \\
    GLEAN & 16to256 & 22.78 & 0.6555 & 21.37 & 0.5456 & 21.56 & 0.6186 \\
    \hline
    Bicubic\_restore & 32to256 & 26.89 & \textbf{0.8252} & 26.65 & \textbf{0.7808} & 27.19 & \textbf{0.8305} \\
    EDSR  & 32to256 & \textbf{28.04} & \underline{0.8182} & \underline{26.87} & 0.7704 & \textbf{27.45} & \underline{0.8208} \\
    RDN   & 32to256 & \underline{27.99} & 0.8164 & 26.77 & 0.7688 & \underline{27.39} & 0.8200 \\
    GLEAN & 32to256 & 27.91 & 0.8173 & \textbf{27.12} & \underline{0.7806} & 27.20 & 0.8141 \\
    \hline
    Bicubic\_restore & 64to256 & 30.75 & 0.8823 & 30.85 & \underline{0.8563} & 30.61 & 0.8784 \\
    EDSR  & 64to256 & 32.00 & 0.8809 & 31.01 & 0.8491 & 31.07 & 0.8726 \\
    RDN   & 64to256 & 31.97 & 0.8803 & 30.92 & 0.8481 & 30.98 & 0.8714 \\
    GLEAN & 64to256 & 27.95 & 0.7964 & 26.97 & 0.7364 & 27.14 & 0.8092 \\
    SRCNN & 64to256 & \textbf{32.03} & \underline{0.8828} & \textbf{31.17} & 0.8526 & \textbf{31.29} & \underline{0.8774} \\
    SRGAN & 64to256 & \underline{32.02} & 0.8814 & \underline{31.13} & 0.8499 & 31.10 & 0.8729 \\
    ESRGAN & 64to256 & 30.68 & 0.8379 & 29.98 & 0.8216 & 29.78 & 0.8298 \\
    Real-ESRGAN & 64to256 & 31.86 & \textbf{0.9001} & 30.52 & \textbf{0.8670} & \underline{31.11} & \textbf{0.8898} \\
    EDVR  & 64to256 & 31.74 & 0.8750 & 30.85 & 0.8423 & 30.97 & 0.8683 \\
    BasicVSR & 64to256 & 29.93 & 0.8343 & 23.99 & 0.6390 & 29.62 & 0.8268 \\
    IconVSR & 64to256 & 30.40 & 0.8495 & 26.34 & 0.7008 & 30.20 & 0.8512 \\
    BasicVSR++ & 64to256 & 30.57 & 0.8434 & 28.49 & 0.7775 & 29.97 & 0.8368 \\
    RealBasicVSR & 64to256 & 26.33 & 0.6444 & 25.25 & 0.6413 & 26.99 & 0.7871 \\
    \hline
    Bicubic\_restore & 128to256 & 35.14 & \underline{0.9313} & 35.03 & \textbf{0.9250} & 33.74 & \textbf{0.9083} \\
    EDSR  & 128to256 & \textbf{36.06} & \textbf{0.9316} & \textbf{35.32} & \underline{0.9194} & \textbf{34.23} & \underline{0.9067} \\
    RDN   & 128to256 & \underline{36.01} & 0.9310 & \underline{35.23} & 0.9188 & \underline{34.19} & 0.9063 \\
    GLEAN & 128to256 & 29.39 & 0.8317 & 28.59 & 0.7895 & 28.22 & 0.8131 \\
    \hline
    \end{tabular}%
  \label{tab:psnr_ssim}%
\end{table*}%

\textcolor{black}{We further corroborated the analysis on the quantitative
of the reconstructed samples in terms of PSNR and SSIM in Table~\ref{tab:psnr_ssim}. Among all the SR methods on the resolution of $64 \times 64$, SRCNN gains the best results of 32.03, 31.17, and 31.29 in terms of PSNR for CASME II, SAMM and SMIC-HS databases, respectively. Real-ESRGAN obtains the best results of 0.9001, 0.9670, and 0.8898 in terms of SSIM for CASME II, SAMM, and SMIC-HS databases, respectively. Additionally, the average PSNRs over ISR methods are 31.22, 30.24, and 30.35 for CASME II, SAMM, and SMIC-HS databases, respectively, while ones over VSR methods 29.79, 26.98, and 29.55 for CASME II, SAMM, and SMIC-HS databases, respectively. On the other hand, the average SSIMs over ISR methods are 0.8657, 0.8321, and 0.8604 for CASME II, SAMM, and SMIC-HS databases, respectively, while ones over VSR methods 0.6093, 0.7202, and 0.8340 for CASME II, SAMM, and SMIC-HS databases, respectively. The experiment results show the ISR method performs significantly better than the VSR method. To compare the results at different resolutions with various SR methods, some SR micro-expression examples are shown in Fig.~\ref{fig:LR_SR_example}. Compared with RealBasicVSR, Real-ESRGAN generated an image closer to the ground truth under~\eg~64to256 case. Qualitative results further demonstrate that ISR methods can generate more real micro-expression face. It is concluded that the existing general ISR methods may be more suitable for MER. }

\textcolor{black}{According to the above-mentioned analysis, if researchers dig into SR methods for micro-expressions, it is suggested that ISR methods would be a good choice with existing universal SR methods and pre-trained models. However, if it is desired to fully utilize the sequence information between micro-expression frames, then exploring effective VSR methods is still necessary. ISR methods can be combined with VSR methods to avoid losing detail information and improve the SR quality of micro-expression sequences.}



\subsection{Experiment Analysis of MER methods}

\subsubsection{\textcolor{black}{Analysis of MER method on all resolutions for all SR methods}}
\label{subsubsec:q1}




\begin{figure}[]
\centering
     \subfigure[]{
       \includegraphics[width=0.5\textwidth]{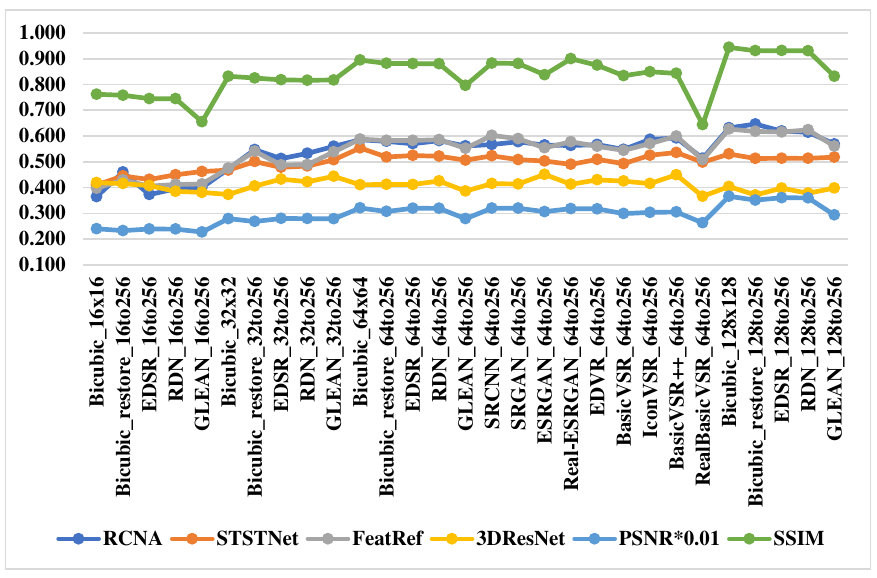}
         \label{fig:imageQ_MERQ_casme2_war}}
     \subfigure[]{
    \includegraphics[width=0.5\textwidth]{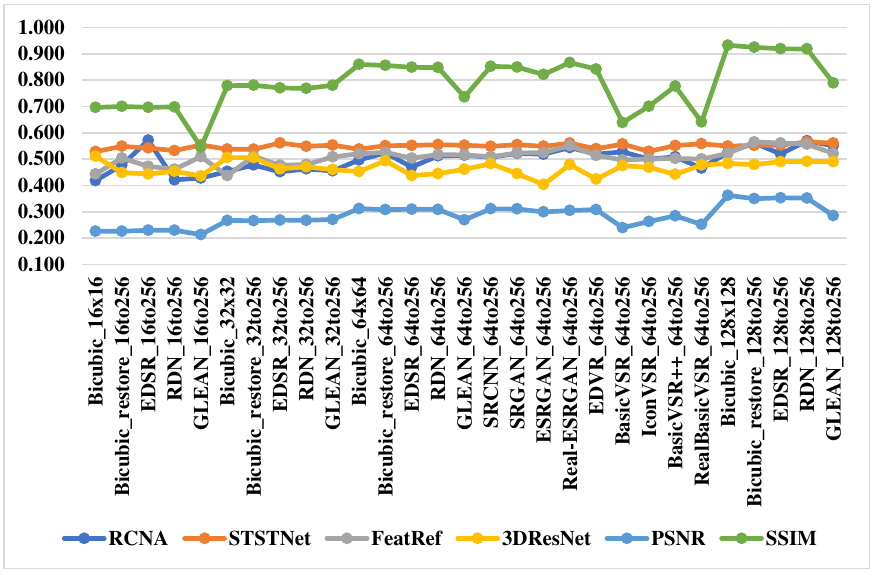}
         \label{fig:imageQ_MERQ_samm_war}
     }
     \subfigure[]{
         \includegraphics[width=0.5\textwidth]{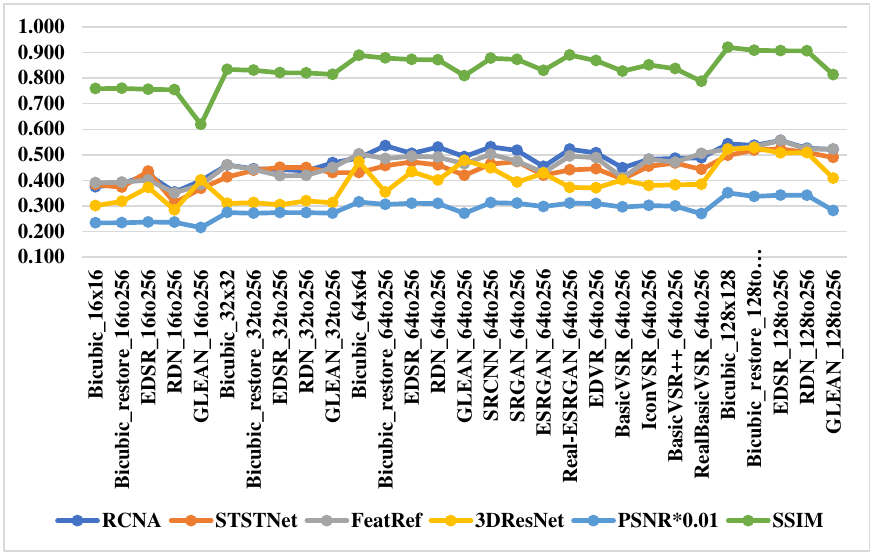}
         \label{fig:imageQ_MERQ_smichs_war}
     }
        \caption{\textcolor{black}{Curve graph of the quality of the reconstructed micro-expression samples in terms of PSNR and SSIM, and MER performance of four methods in terms of WAR on (a) CASME II, (b) SAMM, and (c) SMIC-HS databases. \textcolor{black}{ The horizontal coordinate are correspondence with the ''SR Method" and ``Resolution" in Table \ref{tab:psnr_ssim}, which are named as ''SR Method\_Resolution".} }}
        \label{fig:imageQ_MERQ_war}
        \vspace{-10pt}
\end{figure}
\textcolor{black}{According to Tables~\ref{tab:war_casme2}, \ref{tab:war_samm}, and \ref{tab:war_smichs}, the RCNA method obtains average results of 0.543, 0.503, and 0.481 on CASME II, SAMM, and SMIC-HS databases, respectively. RCNA outperforms FeatRef by 0.001 and 0.014 across all SR methods on CASME II and SMIC-HS databases.~\textcolor{black}{It is noted that RCNA works worse than LIPBSIP on SAMM database. It is due to that SVM is overfitting for LBPSIP}. Indeed, it makes sense to say that the RCNA method outperforms other methods, which is robust to various sample resolutions, because the RCNA method used multiple sample resolutions input in its framework, resulting in well compatibility with different resolutions, especially LR samples. This kind of strategy allows RCNA to achieve better performance compared to other MER methods on average. Additionally, it gives an insight that multi-resolution input may be particularly useful in practical applications where certain databases may only have low-resolution samples available.~\textcolor{black}{On the other hand, except on SAMM database, optical-flow deep learning based MER methods perform significantly better than classical feature engineered methods, and also outperform sequence deep learning based MER methods.} The experiments results indicate that optical flow based deep learning MER methods is a good way to SR strategies.}


\subsubsection{\textcolor{black}{Analysis of MER method on all resolution for a specific SR}} \label{subsubsec:q2}


\textcolor{black}{According to Tables \ref{tab:war_casme2}, \ref{tab:war_samm}, and \ref{tab:war_smichs}, the average performance of each specific MER method under each SR method based on all resolutions is shown in Fig. \ref{fig:avg_war}. Compared to other MER methods, RCNA performs relatively well on all three databases, followed by FeatRef. This is due to the stronger adaptability of RCNA in adapting to different resolution of micro-expressions, and RCNA performs better in some complex situations. Although FeatRef performs slightly worse than RCNA, it still outperforms other methods, possibly due to the specific-expression feature learning  used in FeatRef, giving it good performance in MER. Therefore, further research is needed to explore the performance and reasons behind these methods.}

\subsubsection{\textcolor{black}{Analysis of MER method for a specific strategy and a specific resolution}} \label{subsubsec:specific strategy}

To deeply analyze which MER method can perform well under a specific SR strategy and a specific resolution, we conduct a three-factor analysis \cite{montgomery2017design} of variance on all MER methods across different resolutions and SR methods. Results of three-factor analysis are report in Table \ref{tab:3factors}, where Factor `Res.' is the resolution size with four levels (\eg, 16to256, 32to256, 64to256, and 128to256), Factor `SR' is the SR methods with four levels (4 SR methods), and Factor `MER' is the MER methods with seven levels (7 MER methods).  \textcolor{black}{It is noted that in ANOVA results, levels with the same effect are grouped into the same category. Therefore, levels within the same group have a similar impact on the final experimental results. In the Table \ref{tab:3factors}, according to the calculated differences after ANOVA, the different levels of a factor are divided into four classes, with the number of classes varying depending on the situation. Category ``a" shows the highest level of efficacy and magnitude compared to the other categories. Category ``b", although displaying lower efficacy and magnitude than category ``a", still performs better than the remaining categories. The remaining categories exhibit progressively lower efficacy and magnitude compared to categories ``a" and ``b".}


\begin{table}[th!]
  \centering
  \small
  \caption{Three-factor analysis on three micro-expression databases. ``Res." means ``Resolution". ~\textcolor{black}{Based on the calculated differences from ANOVA, the levels are grouped into a-d categories, with the levels being ordered by the significance of their effects and magnitudes from highest to lowest.}}
    \begin{tabular}{|p{0.75cm}|c|c|c|c|}
    \hline
    Factors & Methods & CASME II & SAMM & SMIC-HS \\
    \hline
   \multirow{5}[4]{*}{Res.} & 16to256  & 0.4084 c & 0.4997 b & 0.3571 c \\
          & 32to256   & 0.4448 b & 0.5093 b  & 0.3699 c \\
          & 64to256   & 0.4841 a & 0.5143 b  & 0.4097 b  \\
          & 128to256  & \textbf{0.4939 a} & \textbf{0.5389 a} & \textbf{0.4467 a} \\
\cline{2-5}          & $p$ value & 5.21E-07 & 3.7459E-05 & 7.91E-07\\
    \hline
    \multirow{6}[4]{*}{SR}
        & Bicubic\_restore & \textbf{0.4630} & \textbf{0.5220} & 0.3978 \\
          & EDSR  & 0.4600 & 0.5179 & \textbf{0.4081}\\
          & RDN   & 0.4600 & 0.5159 & 0.3905 \\
          & GLEAN & 0.4630 & 0.5171 & 0.3972 \\
\cline{2-5}          & $p$ value & 0.3267 & 0.2944 & 0.3848\\
    \hline
    \multirow{8}[4]{*}{MER} & LBPTOP & 0.4413 c & 0.5651 ab & 0.3478 d\\
          & LBPSIP & 0.4593 c & \textbf{0.572 a}  & 0.3310 d \\
          & 3DHOG & 0.3513 e  & 0.4520 e & 0.3256 d\\
          & RCNA  & \textbf{0.5291 a} & 0.4933 c & \textbf{0.4713 a}\\
          & STSTNet & 0.4929 b & 0.5489 b & 0.4420 b\\
          & FeatRef & 0.5264 a & 0.50605 c & 0.4603 ab\\
          & 3DResNet & 0.4043 d & 0.4715 d & 0.3927 c\\
\cline{2-5}   & $p$ value & 3.59E-09 & 5.11E-09 & 1.75E-08 \\
    \hline
    \end{tabular}%
  \label{tab:3factors}%
\end{table}%

\textcolor{black}{Firstly, it is can be seen from Table \ref{tab:3factors} that for the resolution size (Factor `Res.'), there is significant difference between the resolution of 128to256 among all the databases, as the resolution of 128to256 are grouped into category of ``a".  It is also verified the conclusion in Section~\ref{subsec:low-res_effect} that LR can indeed cause a significant decrease in performance.}
Furthermore, in Factor `SR', it is seen that the p-values are 0.3267 ($p>0.05$), 0.2944 ($p>0.05$), and 0.3848 ($p>0.05$) on the CASME II, SAMM, and SMIC-HS databases, respectively. It indicates that there is no significant difference between different SR methods.
\textcolor{black}{Finally, similar to  the the conclusion in \ref{subsubsec:q1}, on Factor `MER', it is reveal that on CASME II database, RCNA and FeatRef are grouped into ``a" class; on SAMM database, LBPTOP and LBPSIP are grouped into ``a" class; and on SMIC-HS, RCNA and FeatRef are grouped into ``a" class. It is revealed that except on SAMM database, as optical flow can capture subtle information about
the changes occurring in micro-expressions, thus RCNA and FeatRef perform better than other methods. The reason for why LBPTOP and LBPSIP works better than other MER methods mainly because that SVM is overfitting for them. }

\textcolor{black}{Therefore, according to the above analysis, the best effect is achieved at the resolution of 128to256 on the three databases. Specifically, on CASME II,  the combination of Bicubic\_restore and RCNA shows the most significant effect; on database SAMM, the combination of Bicubic\_restore with  LBPSIP shows the most significant effect; and in database SMIC-HS, the combination of EDSR  with  RCNA works best.
These conclusions also corroborate with the results shown in Tables \ref{tab:war_casme2}, \ref{tab:war_samm}, and \ref{tab:war_smichs}.
Except the overfitting occurs on SAMM, the best results   achieved on CASME II and SAMM are based on the combination of resolution of 128to256 (specific resolution), Bicubic\_restore (specific strategy ) and RCNA (MER method).  Those phenomenon further demonstrates  the effectiveness of the optical flow-based deep learning method for MER. Also,  some key points can be achieved. Firstly, collecting high resolution micro-expression samples is essential, as the better MER results are got on higher resolution of $128 \times 128$, other than other low resolutions. Secondly, with a higher resolution, even the simple  Bicubic\_restore method may performs  better than some general super-resolution methods on micro-expression samples. This indicates that  it is indeed necessary to study specific SR methods for MER. }

\subsection{The correlation of SR and MER methods}
\label{subsec:q5}

\textcolor{black}{In order to analyze the influence of image restoration quality to MER, the SSIM and PSNR of SR and the performance of MER are shown in Fig. \ref{fig:imageQ_MERQ_war}. From the perspective of evaluating the quality of sample restoration, SR methods with higher peak PSNR and SSIM have better image restoration quality. According to Fig.~\ref{fig:imageQ_MERQ_war}, when these high-quality restored samples are used as inputs, MER methods gain the considerable performance. The results suggest that if we want to develop an end-to-end SR-MER model for LR MER, sample restoration quality may be jointly taken into consideration in loss function, thereby improving the model performance. It may have the following advantages:}
   (I) The sample restoration quality loss may be simultaneously optimized with MER task loss during the training process, allowing the model to learn the right image quality for MER. This is different from independently completing image restoration before sending it to the MER model. 
    (II) Jointly optimizing the multi-task losses can mutually promote and constrain each other, making the model proficient in both high-quality image restoration and MER, achieving twice the result with half the effort.
   (III) End-to-end training consolidates the relevant knowledge of both image restoration and MER tasks, allowing the network to have a deeper understanding of their inherent connections and dependencies, and a stronger learning and understanding ability for the overall task.


\subsection{Discussion}

Our benchmarking and comprehensive study aims to provide valuable reference for researchers, establish a data comparison baseline for subsequent LR MER, and also guide the development of new  methods suitable for LR MER. According to the experimental analysis, we attempt to find the suitable strategy for the THREE questions raised in Section~\ref{sec:Intro}. Noted that there is no one-size-fits-all answer to these questions, as SR-MER method relies on various factors such as the specific SR strategy being used and the type of data being processed.

\textbf{Q1: Among those evaluated MER methods, which kind is the best for all SR strategies or a specific SR strategy across all resolutions? }  

According to the analysis in Section \ref{subsubsec:q1} and Section~\ref{subsubsec:q2},
comparing the different type of MER methods,  optical-flow deep learning based MER
methods perform  superior to classical feature engineered methods, and also outperform sequence deep learning
based MER methods. It is mainly due to the following reasons: (I) Optical flow can capture more information about the changes occurring in micro-expressions and describe the motion patterns between pixels, which can more accurately represent facial muscle distortions and eye movements in micro-expressions.
(II) Optical flow features combined with deep learning can learn more effective representations of micro-expression changes. In contrast, handcrafted features lack generalization ability and optimization space.

For those specific MER methods, RCNA achieves the comparable better performance than other methods. Moreover, for a specific SR strategy, RCNA also perform well across all resolutions. It is mainly due to that RCNA method was designed to be compatible with inputs of multiple sample resolutions, especially low resolution samples.  

In summary, when designing a MER method that is compatible with multi-resolution inputs, an optical flow-based method is a good choice. Moreover, considering compatibility with multiple resolutions during design will be helpful for improving performance.





 \textbf{Q2: From the perspective of micro-expression samples, is video SR methods better than image SR methods?}

In Section \ref{subsec:Analysis of SR methods}, experiment results indicate that ISR methods sounds more suitable for MER than VSR methods. ISR methods mainly focus on super-resolution reconstruction of a single image. These methods typically generate a higher resolution image by learning the texture and detail information of a high-resolution image based on an input image. However, due to the brief nature of micro-expressions, a single image may involve more sufficient dynamic information to describe the complete micro-expression information especially the sequence pattern.
For example, in the case of ``GLEAN\_16to256" in Fig.~\ref{fig:LR_SR_example},  due to the inability to capture the relationships between consecutive samples, ISR methods may result in significant differences among images in the same sequence after SR, which is not conducive to subsequent use of sequence-based MER methods.
 Therefore, in the future, it is recommended to incorporate information between micro-expression sequences into the generalized ISR methods to develop specific SR methods for micro-expressions.

\textbf{Q3: Whether there is a proportional relationship between the quality of the super-resolved samples and the recognition results in micro-expression SR samples?}

With the analysis in Section \ref{subsec:q5} and Fig. \ref{fig:imageQ_MERQ_war}, it is concluded that high-quality super-resolved samples can lead to better MER performance, and there is a proportional relationship between the two. It means that improving the quality of super-resolved samples can enhance the accuracy of MER. Thus, generating high-quality micro-expression SR samples is crucial for improving MER accuracy. This can be achieved by selecting appropriate SR algorithms, optimizing the training process, adding more constraints, using higher resolution original images, and combining other techniques such as image denoising, enhancement, and restoration.


\section{CONCLUSION}

In this paper, we have extensively investigated the SR-MER problem, performing a comprehensive benchmark evaluation from two distinct perspectives: SR and the use of MER methods for LR MER. We raised four questions that we hope to answer through the benchmark. Our investigation centers around addressing four fundamental questions, which we aimed to elucidate through this benchmark. Guided by these questions, we meticulously devised the entire benchmark workflow. Initially, our experiments are founded on three databases: CASME II, SAMM, and SMIC-HS. After extracting facial regions from the original samples, we standardized their sizes and subsequently down-scaled them using the Bicubic method. Subsequently, we gathered a collection of 13 renowned SR techniques and 7 widely employed MER methods. We engaged in super-resolution on samples sourced from the aforementioned databases and then conducted MER analyses based on the super-restored samples. Alongside micro-expression analysis, we performed an in-depth assessment of the quality of super-resolution in terms of PNSR and SSIM. Moreover, according to the initial experimental findings, statistical analyses, multi-factor investigations, and curve graph assessments, we successfully addressed the series of core questions we initially posed. Consequently, we not only clarified our research inquiries but also illuminated potential directions for future studies. In conclusion, we aspire for our work to foster advancements in micro-expression analysis and to inspire researchers to delve into the intricacies of the SR-MER challenge. Going forward, our efforts will persist in accumulating a broader array of effective end-to-end SR-MER methodologies tailored for micro-expressions.


\footnotesize
\bibliographystyle{IEEEtran}
\bibliography{ref_tmm}

\begin{thebibliography}{10}
\providecommand{\url}[1]{#1}
\csname url@samestyle\endcsname
\providecommand{\newblock}{\relax}
\providecommand{\bibinfo}[2]{#2}
\providecommand{\BIBentrySTDinterwordspacing}{\spaceskip=0pt\relax}
\providecommand{\BIBentryALTinterwordstretchfactor}{4}
\providecommand{\BIBentryALTinterwordspacing}{\spaceskip=\fontdimen2\font plus
\BIBentryALTinterwordstretchfactor\fontdimen3\font minus \fontdimen4\font\relax}
\providecommand{\BIBforeignlanguage}[2]{{%
\expandafter\ifx\csname l@#1\endcsname\relax
\typeout{** WARNING: IEEEtran.bst: No hyphenation pattern has been}%
\typeout{** loaded for the language `#1'. Using the pattern for}%
\typeout{** the default language instead.}%
\else
\language=\csname l@#1\endcsname
\fi
#2}}
\providecommand{\BIBdecl}{\relax}
\BIBdecl

\bibitem{Zhang2018a}
F.~Zhang, T.~Zhang, Q.~Mao, and C.~Xu, ``Joint pose and expression modeling for facial expression recognition,'' in \emph{Proceedings of CVPR}, 2018, pp. 3359--3368.

\bibitem{Ekman1969}
P.~Ekman and W.~V. Friesen, ``Nonverbal leakage and clues to deception?'' \emph{Psychiatry-interpersonal \& Biological Processes}, vol.~32, no.~1, pp. 88--106, 1969.

\bibitem{Ekm2009}
P.~Ekman, \emph{Telling Lies: Clues to Deceit in the Marketplace, Politics,and Marriage (Revised Edition)}.\hskip 1em plus 0.5em minus 0.4em\relax WW Norton \& Company, 2009.

\bibitem{Michael2010}
N.~Michael, M.~Dilsizian, D.~N. Metaxas, and J.~K. Burgoon, ``Motion profiles for deception detection using visual cues,'' in \emph{Proceedings of ECCV}, 2010, pp. 462--475.

\bibitem{Liong2018}
S.~Liong, J.~See, R.~C. Phan, K.~Wong, and S.~Tan, ``Hybrid facial regions extraction for micro-expression recognition system,'' \emph{Journal of Signal Processing Systems}, vol.~90, no.~4, pp. 601--617, 2018.

\bibitem{Ngo2017}
A.~C.~L. Ngo, J.~See, and R.~C. Phan, ``Sparsity in dynamics of spontaneous subtle emotions: Analysis and application,'' \emph{{IEEE} Trans. Affect. Comput.}, vol.~8, no.~3, pp. 396--411, 2017.

\bibitem{Zhao2007}
G.~Zhao and M.~Pietik{\"{a}}inen, ``Dynamic texture recognition using local binary patterns with an application to facial expressions,'' \emph{IEEE Trans. Pattern Anal. Mach. Intell.}, vol.~29, no.~6, pp. 915--928, 2007.

\bibitem{Huang2015}
X.~Huang, S.~Wang, G.~Zhao, and M.~Pietik{\"{a}}inen, ``Facial micro-expression recognition using spatiotemporal local binary pattern with integral projection,'' in \emph{Proceedings of ICCV}, 2015, pp. 1--9.

\bibitem{Huang2016}
X.~Huang, G.~Zhao, X.~Hong, W.~Zheng, and M.~Pietik{\"{a}}inen, ``Spontaneous facial micro-expression analysis using spatiotemporal completed local quantized patterns,'' \emph{Neurocomputing}, vol. 175, pp. 564--578, 2016.

\bibitem{Huang2019}
X.~Huang, S.~Wang, X.~Liu, G.~Zhao, X.~Feng, and M.~Pietik{\"{a}}inen, ``Discriminative spatiotemporal local binary pattern with revisited integral projection for spontaneous facial micro-expression recognition,'' \emph{{IEEE} Trans. Affect. Comput.}, vol.~10, no.~1, pp. 32--47, 2019.

\bibitem{Xia2020}
Z.~Xia, X.~Hong, X.~Gao, X.~Feng, and G.~Zhao, ``Spatiotemporal recurrent convolutional networks for recognizing spontaneous micro-expressions,'' \emph{{IEEE} Trans. Multi.}, vol.~22, no.~3, pp. 626--640, 2020.

\bibitem{Gan2019}
Y.~S. Gan, S.~Liong, W.~Yau, Y.~Huang, and T.~L. Ken, ``Off-apexnet on micro-expression recognition system,'' \emph{Signal Process. Image Commun.}, vol.~74, pp. 129--139, 2019.

\bibitem{Zhou2019}
L.~Zhou, Q.~Mao, and L.~Xue, ``Dual-inception network for cross-database micro-expression recognition,'' in \emph{Proceedings of FG}, 2019, pp. 1--5.

\bibitem{Mao2022}
Q.~Mao, L.~Zhou, W.~Zheng, X.~Shao, and X.~Huang, ``Objective class-based micro-expression recognition under partial occlusion via region-inspired relation reasoning network,'' \emph{{IEEE} Trans. Affect. Comput.}, vol.~13, no.~4, pp. 1998--2016, 2022.

\bibitem{Zhou2022}
L.~Zhou, Q.~Mao, X.~Huang, F.~Zhang, and Z.~Zhang, ``Feature refinement: An expression-specific feature learning and fusion method for micro-expression recognition,'' \emph{Pattern Recognit.}, vol. 122, p. 108275, 2022.

\bibitem{Li2019c}
G.~Li, J.~Shi, J.~Peng, and G.~Zhao, ``Micro-expression recognition under low-resolution cases,'' in \emph{Proceedings of VISIGRAPP}, 2019, pp. 427--434.

\bibitem{Sharma2022}
P.~Sharma, S.~Coleman, P.~Yogarajah, L.~Taggart, and P.~Samarasinghe, ``Comparative analysis of super-resolution reconstructed images for micro-expression recognition,'' \emph{Advances in Computational Intelligence}, vol.~2, p.~24, 2022.

\bibitem{Sharma2022a}
P.~Sharma, S.~Coleman, P.~Yogarajah, L.~Taggart, and P.~Samarasinghe, ``Evaluation of generative adversarial network generated super resolution images for micro expression recognition,'' in \emph{Proceedings of ICPRAM}, 2022, pp. 560--569.

\bibitem{Shi2018}
J.~Shi, X.~Liu, Y.~Zong, C.~Qi, and G.~Zhao, ``Hallucinating face image by regularization models in high-resolution feature space,'' \emph{{IEEE} Trans. Image Process.}, vol.~27, no.~6, pp. 2980--2995, 2018.

\bibitem{Wang2018a}
X.~Wang, K.~Yu, S.~Wu, J.~Gu, Y.~Liu, C.~Dong, Y.~Qiao, and C.~Change~Loy, ``{ESRGAN}: Enhanced super-resolution generative adversarial networks,'' in \emph{Proceedings of ECCVW}, 2018, pp. 63--79.

\bibitem{Rakotonirina2020}
N.~C. Rakotonirina and A.~Rasoanaivo, ``{ESRGAN+} : Further improving enhanced super-resolution generative adversarial network,'' in \emph{Proceedings of ICASSP}, 2020, pp. 3637--3641.

\bibitem{Ma2020}
C.~Ma, Z.~Jiang, Y.~Rao, J.~Lu, and J.~Zhou, ``Deep face super-resolution with iterative collaboration between attentive recovery and landmark estimation,'' in \emph{Proceedings of CVPR}, 2020, pp. 5569--5578.

\bibitem{Chan2021}
K.~C. Chan, X.~Wang, X.~Xu, J.~Gu, and C.~C. Loy, ``{GLEAN}: Generative latent bank for large-factor image super-resolution,'' in \emph{Proceedings of CVPR}, 2021, pp. 14\,240--14\,249.

\bibitem{Wang2021a}
X.~Wang, L.~Xie, C.~Dong, and Y.~Shan, ``Real-{ESRGAN}: Training real-world blind super-resolution with pure synthetic data,'' in \emph{Proceedings of ICCV}, 2021, pp. 1905--1914.

\bibitem{Zhang2018b}
Y.~Zhang, Y.~Tian, Y.~Kong, B.~Zhong, and Y.~Fu, ``Residual dense network for image super-resolution,'' in \emph{Proceedings of CVPR}, 2018, pp. 2472--2481.

\bibitem{Wang2019b}
X.~Wang, K.~C. Chan, K.~Yu, C.~Dong, and C.~C. Loy, ``{EDVR}: Video restoration with enhanced deformable convolutional networks,'' in \emph{Proceedings of CVPRW}, June 2019, pp. 1954--1963.

\bibitem{Chan2022}
K.~C. Chan, S.~Zhou, X.~Xu, and C.~C. Loy, ``Basic{VSR}++: Improving video super-resolution with enhanced propagation and alignment,'' in \emph{Proceedings of CVPR}, 2022, pp. 5972--5981.

\bibitem{Chan2022a}
K.~C. Chan, S.~Zhou, X.~Xu, and C.~C. Loy, ``Investigating tradeoffs in real-world video super-resolution,'' in \emph{Proceedings of CVPR}, 2022, pp. 5962--5971.

\bibitem{Li2013}
X.~Li, T.~Pfister, X.~Huang, G.~Zhao, and M.~Pietik{\"{a}}inen, ``A spontaneous micro-expression database: Inducement, collection and baseline,'' in \emph{Proceedings of FG}, 2013, pp. 1--6.

\bibitem{Yan2014}
W.~Yan, X.~Li, S.~Wang, G.~Zhao, Y.~Liu, Y.~Chen, and X.~Fu, ``{CASME II}: An improved spontaneous micro-expression database and the baseline evaluation,'' \emph{PLOS ONE}, vol.~9, no.~1, pp. 1--8, 2014.

\bibitem{Davison2018}
A.~K. Davison, C.~Lansley, N.~Costen, K.~Tan, and M.~H. Yap, ``{SAMM:} {A} spontaneous micro-facial movement dataset,'' \emph{{IEEE} Trans. Affect. Comput.}, vol.~9, no.~1, pp. 116--129, 2018.

\bibitem{Liong2019a}
S.~Liong, Y.~S. Gan, J.~See, and H.~Khor, ``A shallow triple stream three-dimensional {CNN} ({STSTNet}) for micro-expression recognition system,'' \emph{CoRR}, vol. abs/1902.03634, 2019.

\bibitem{Ledig2016}
C.~Ledig, L.~Theis, F.~Husz{\'a}r, J.~Caballero, A.~Cunningham, A.~Acosta, A.~Aitken, A.~Tejani, J.~Totz, and Z.~Wang, ``Photo-realistic single image super-resolution using a generative adversarial network,'' in \emph{Proceedings of CVPR}, 2016, pp. 4681--4690.

\bibitem{Dong2016}
C.~Dong, C.~C. Loy, K.~He, and X.~Tang, ``Image super-resolution using deep convolutional networks,'' \emph{{IEEE} Trans. Pattern Anal. Mach. Intell.}, vol.~38, no.~2, pp. 295--307, 2016.

\bibitem{Lim2017}
B.~Lim, S.~Son, H.~Kim, S.~Nah, and K.~Mu~Lee, ``Enhanced deep residual networks for single image super-resolution,'' in \emph{Proceedings of CVPR}, 2017, pp. 136--144.

\bibitem{Kelvin2020}
K.~C.~K. Chan, X.~Wang, K.~Yu, C.~Dong, and C.~C. Loy, ``Basicvsr: The search for essential components in video super-resolution and beyond,'' \emph{CoRR}, vol. abs/2012.02181, 2020.

\bibitem{Wang2014}
S.~Wang, W.~Yan, X.~Li, G.~Zhao, and X.~Fu, ``Micro-expression recognition using dynamic textures on tensor independent color space,'' in \emph{Proceedings of ICPR}, 2014, pp. 4678--4683.

\bibitem{Polikovsky2009}
S.~Polikovsky, Y.~Kameda, and Y.~Ohta, ``Facial micro-expressions recognition using high speed camera and 3d-gradient descriptor,'' in \emph{Proceedings of ICDP}, 2009, pp. 1--6.

\bibitem{Xia2020a}
Z.~Xia, W.~Peng, H.~Khor, X.~Feng, and G.~Zhao, ``Revealing the invisible with model and data shrinking for composite-database micro-expression recognition,'' \emph{{IEEE} Trans. Image Process.}, vol.~29, pp. 8590--8605, 2020.

\bibitem{Hara2017}
K.~Hara, H.~Kataoka, and Y.~Satoh, ``Learning spatio-temporal features with 3d residual networks for action recognition,'' in \emph{Proceedings of ICCV}, 2017, pp. 3154--3160.

\bibitem{He2016a}
K.~He, X.~Zhang, S.~Ren, and J.~Sun, ``Deep residual learning for image recognition,'' in \emph{Proceedings of CVPR}, 2016, pp. 770--778.

\bibitem{montgomery2017design}
D.~C. Montgomery, \emph{Design and Analysis of Experiments}.\hskip 1em plus 0.5em minus 0.4em\relax Wiley, 2017.

\end{thebibliography}

\end{document}